REVIEW

# Emergence of Agentic AI: A Review on Evolution, Background, Working Principles, Applications, Adoption Factors, and Future Research Directions

**AKM Bahalul Haque[1,*], Al Amin Islam Ridoy[2], Mohammad Rayhan[3] and Ivan Porres[1]**

[1]Åbo Akademi University, Turku, Finland
[2]Rajshahi University of Engineering and Technology, Rajshahi, Bangladesh
[3]LUT University, Lappeenranta, Finland
*Corresponding Author: AKM Bahalul Haque. Email: akmbahalul.haque@abo.fi

**ABSTRACT:** Agentic AI is gaining new insights and advancements in the field of Artificial Intelligence, fostering significant potential to enable rapid transformation across various domains. This rapid advancement and the potential to revolutionize various domains advocate the need for a deeper understanding and firm grasp of the technology. Moreover, an investigation into state-of-the-art research directions in agentic AI should be conducted to comprehensively assess the potential scope for improvement and application. Therefore, to address these objectives, a comprehensive review can provide researchers and practitioners with valuable insights into the current state and future research scopes of agentic AI. Hence, this work considers the recently published scholarly contributions in agentic AI across various domains and (i) discusses the fundamentals and working principles of Agentic AI, (ii) traces the historical and theoretical evolution of agency in artificial systems, (iii) explores and discusses Agentic AI's architecture, working principles, and functionalities, (iv) explores real-world applications of Agentic AI across various domains, (v) analyzes the research findings, identifies current challenges, and discuss potential future research directions, and (vi) proposes a comprehensive framework of stakeholders' intention to use and adopt Agentic AI with the help of proposed system quality dimensions. Therefore, this systematic review provides researchers and practitioners with a comprehensive understanding of Agentic AI, its current developments and applications, highlights key research gaps, and outlines future research directions.

## 1 Introduction

Artificial Intelligence (AI) is undergoing rapid transformation, revolutionizing various sectors of our daily lives [1]. The transformation ranges from rule-based systems to personalized recommendation systems and human-like responsive systems [2]. As a result of this transformation, agentic artificial intelligence (Agentic AI) has been introduced as a system that is autonomous, independent, and goal-oriented, aiming to supplement the functionality of the future AI landscape [3]. Agentic AI systems react to specific inputs and perceive environmental contexts that guide their actions. Traditional AI systems operate with fixed-term goals, use task-specific models, and use a human-in-the-loop architecture. On the contrary, an Agentic AI system's proactive and reactive nature is channeled through contextual actions and reasoning [4]. Agentic AI uses the concepts of multi-agent, autonomous, and distributed systems.

Agentic AI, characterized by adaptability, autonomy, and goal-oriented behavior, has evolved from rule-based systems to today's hierarchical, multi-agent, collaborative framework. Some early breakthroughs of Agentic AI can be seen after the evolution of AutoGPT and AutoML. Agentic AI divides complex tasks into sub-tasks and simplifies them. Later, these sub-tasks are automated and require minimal human interaction [5]. The emergence of Agentic AI has applications in healthcare, customer support, banking, e-commerce, social media, and transportation [6]. In the healthcare ecosystem, Agentic AI can monitor medical data, diagnostic information, and anomalies. The financial domain utilizes intelligent bots with advanced predictive models, while the education field employs agentic systems for tutoring, collecting, and analyzing feedback and pupil engagement statistics [7]. Similarly, this technology can be integrated into the Internet of Things (IoT) ecosystem for smart home management, autonomous agricultural systems, wearable devices, smart buildings, and more. Agentic capabilities are also used in software engineering, improving business process management and generating adaptive workflows [8].

The agile transformation of Agentic AI can improve productivity or innovation, but it can also raise concerns in terms of computational costs, transparency, fairness, privacy, security, and ethics. Besides, there are other limitations, such as data unavailability or integration mismatches [9]. Recent scholarly contributions have discussed various perspectives of Agentic AI, such as being an active participant and leveraging a sociotechnical lens to personalize services and streamline resource allocation [10]. The transformative potential of Agentic AI is also explored in the financial technology sector (FinTech), where 85% of people's opinions on social media were enthusiastic about adopting Agentic AI [11]. The role of Agentic AI extends to biomedical research, particularly in quantitative clinical pharmacology, harnessing agentic workflows to enhance efficacy, but privacy and computational demands remain hurdles [12]. Scalable tools like LangGraph and AutoGen offer solutions, yet their success depends on addressing ethical and security barriers, ensuring Agentic AI's transformative impact integrates seamlessly across every applied sector of AI. Similarly, another recent literature review also discusses the characteristics, applications, and future potential of Agentic AI in business and society [13]. However, to the best of our knowledge, no comprehensive review has been published that discusses the background, working principles, applications, benefits, research gaps, and future research potential of Agentic AI.

The research gaps mentioned above highlight the necessity for scholarly contributions that thoroughly examine the architecture, working principles, historical evolution, applications across various domains, research deficiencies, and future research directions. Therefore, this work aims to address the research gaps identified in the current literature by fulfilling the following research objectives:

- Comprehensively discuss and trace the evolution and attributes of agency in artificial systems.
- Analyze and discuss the layered architecture, different components, and functions of Agentic AI.
- Explore and discuss the real-world applications of Agentic AI across various domains based on recent scholarly contributions.
- Identify the adoption factors of Agentic AI, which are also proposed as system quality dimensions of Agentic AI systems.
- Critically analyze the research findings, identify current challenges, and problematize potential future research directions.
- Propose a comprehensive framework for stakeholders' intention to use and adopt Agentic AI, utilizing the proposed system quality dimensions.

The rest of the paper is organised as: Section 2 contains the Research strategy, Section 3 discusses the evolutionary background of Agentic AI, Section 4 outlines the architecture and workflow of Agentic AI, Section 5 identifies the real-life applications of Agentic AI, Section 6 synthesizes the system quality dimensions (adoption factors of Agentic AI usage), Section 7 discusses the proposed synthesized framework for Agentic AI adoption and use, Section 8 outlines the research gaps and future research directions, contributions of this work are discussed in Section 9, limitation of this work is discussed in Section 10 followed by the conclusion in Section 11.

### *Motivation*

The ambition to develop machines with human-like cognition and advanced problem-solving capability, where systems can learn from data, carry out complex reasoning, continuously adapt to new situations, and take action or make decisions autonomously has been the core driving force behind the research in artificial intelligence (AI) [14]. Despite the remarkable advancements of the AI domain, traditional approaches struggled to achieve these objectives due to inherent constraints in their architectures and operational paradigms. Only Agentic AI, with its autonomy in goal-directed behavior, reasoning, adaptability, planning, and the ability to take actions, offers an expedient way to achieve these goals.

The pursuit of autonomous intelligence has been shaped by the realization that intelligence extends beyond the ability to process information, recognize patterns, or generate content, but rather depends on the ability to initiate goal-directed behavior in complex, dynamic environments. This realization compels us to acknowledge that the recent AI architectures' confinement to reactive responses is insufficient due to their dependency on human intervention, limited adaptability to novel situations, and inability to pursue complex, multi-step objectives independently [15]. So, next-generation systems must become proactive by interpreting high-level objectives, constructing multi-stage plans, and executing operations independently to solve complex problems with minimal human supervision.

## 2 Research Strategy

In this work, we have adopted a systematic literature review technique for software engineering as outlined in [16] [17]. Figure 1 illustrates the chronological process of article search and the selection strategy used in this work. As the figure shows, we first identified suitable search keywords to facilitate database searches. Our target is to conduct a systematic review of Agentic AI, including its evolution, background, and working principles. The term "AI Agent" is often used interchangeably with Agentic AI, though there are differences in architecture, working principles, scope of use, and deployment. Therefore, for search keywords, we will be using Agentic AI and AI Agents. While screening, we will remove articles that discuss AI Agents exclusively. The following search string will be used for the search:

*( "Agentic AI" OR "Intelligent agent" OR "AI Agents" OR "Agentic artificial intelligence" OR "Agentic LLM" OR "Agentic system" OR "AI autonomy" OR "Autonomous decision-making" OR "Goal-driven AI" OR "Self-directed AI" OR "Self-adaptive agent" )* **AND** *( "Adoption Factors" OR "Application" OR "System Quality" OR "Working Principles" )*

The Scopus and Web of Science databases were selected as search databases because both are widely recognized for their comprehensive indexing of scholarly publications. After that, the inclusion and exclusion criteria were used to screen and identify the articles for review. To maintain methodological rigor, a structured quality assessment was conducted during the review process. For the quality assessment, inclusion and exclusion criteria have been used. Moreover, each article was evaluated across multiple dimensions, including methodological soundness, validity, and relevance to the research questions. The authors independently assessed all included studies. Later, each selection was discussed together, and disagreements were resolved through brainstorming and reflective discussion. Each author presented their justification during the discussion, and upon reaching agreement, the articles were included in the final list. The inclusion and exclusion criteria are as follows:

**Inclusion criteria**

- Articles published in journals and conference proceedings are included.
- Articles published between 2021 and 2025 are included.
- Articles that focus on conceptual or experimental implementation are included.

**Exclusion criteria**

- Articles published other than in English are excluded from the list.
- Review articles, book chapters, editorials, and other types of articles are excluded from consideration.
- Unavailable full-text articles are excluded.

Initially, the "search keywords" were used to search within the article's "title, abstract, and keywords" in the Scopus and Web of Science databases. Searches were limited to the years 2021–2025. The search returned 3331 records from Scopus and 799 from Web of Science. Later, in accordance with the inclusion and exclusion criteria, only journals and conference proceedings published in English were selected. The screening resulted in 2492 records from Scopus and 253 records from Web of Science. Then, both databases were combined, and duplicate articles were removed.

A total of 2608 articles were screened for inclusion and exclusion based on the criteria, with titles and abstracts reviewed. The screening resulted in 593 studies. These studies were then further screened by reading and skimming through the full text. 72 articles remained after the screening. Later, to understand and discuss the history, evolution, and architecture of Agentic AI, we further selected 22 articles through snowball sampling. In total, 94 articles have been used to discuss the research objectives of this paper.

## 3 Evolutionary Background of Agentic AI

The evolution of artificial intelligence represents a profound transformation from a rule-based, rigid system to a flexible, self-organizing cognitive network. In the early days, symbolic AI (1950s–1970s) was one of the intelligent approaches, in which researchers encoded all knowledge by hand; back then, intelligence was conceived as applying fixed rules within a closed environment [18]. However, as the complexity of real-world data and the need for more intelligent devices increased, such deterministic models gradually gave way to the statistical learning (1980s–2010s) era, when emphasis was given on pattern discovery from datasets rather than depending on handcrafted rules [19,20]. This paradigm proved that intelligence could emerge from statistical data analysis, probabilistic reasoning, or pattern-based learning. The subsequent transition from statistical learning toward generative AI (2010s–2020s) accelerated when AI models began producing creative content rather than focusing on interpreting or analyzing inputs [21,22]. However, isolation from the real world, limited interaction, and the inability to take independent action/decision for system or environment controlling gave rise to the need for the next AI Agent paradigm [23].

AI agent (2020s–present) bridges the gap between content generation and real-world interaction. This paradigm represents a significant shift as systems evolve from isolated computational units toward interconnected cognitive ecosystems. The advanced tools feature augmented the autonomy of AI agents through LLM-API frameworks, enabling autonomous reasoning, planning, and complex task execution [24]. Yet, its autonomy is limited by narrow context awareness, task-specific restrictions, and a hard-coded tool determination process. That's why the evolution escalated towards Agentic AI (emerging), which is driven by multi-agent orchestration, huge context awareness through cross-agent coordination, where agents play specialized roles to solve complex problems through distributed collaborative intelligence [25,26]. The complete trajectory of AI evolution is presented in Table 1, and it shows how AI has evolved from merely replicating human cognition towards an entirely new set of virtual collective intelligence through Agentic AI.

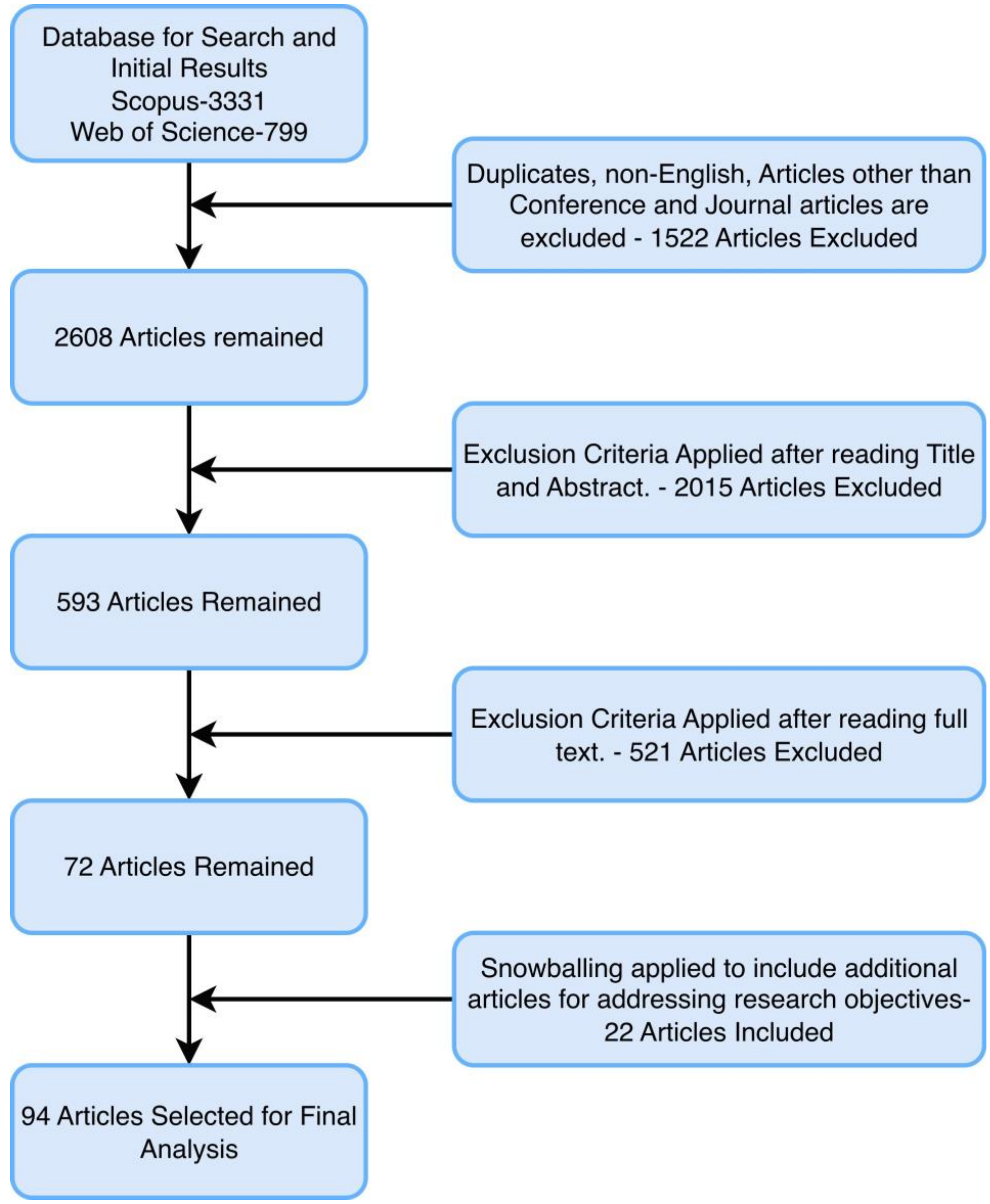

**Figure 1:** Article Search and Selection Process

**Table 1:** The transition from traditional reactive AI systems to proactive Agentic AI systems.

| Era | Time frame | Paradigm | Key technology | Limitations | How transition happened |
|---|---|---|---|---|---|
| Symbolic AI [18] | 1950s to 1970s | Rules as intelligence | Logic programming | • Handwritten knowledge—no perceptual grounding.<br>• No learning or adaptation.<br>• Cannot handle the slightest change in input. | • **Upgrade:** Data-driven models and pattern recognition.<br>• **Mechanism:** Statistical methods (ML) and Deep Learning (DL) architectures.<br>• **Impact:** Replaced manual rule-coding, enabled autonomous data-driven learning, and can handle variability. |
| Statistical Learning [19,20] | 1980s to 2010s | Pattern recognition | ML algorithms, Backpropagation | • Requires massive labeled data.<br>• Poor generalization; false responses outside training data.<br>• Needs manual updates for new data/environment patterns. | • **Upgrade:** Self-supervised generative models with few/one/zero-shot capabilities.<br>• **Mechanism:** Transformers (Self-attention) trained on massive unlabeled data (parallel GPUs/TPUs).<br>• **Impact:** No need for labeled data, enables cross-context reasoning and reuse without retraining. |
| Gen AI [21,22] | 2010s to present | Content generation | GANs, Transformers | • Hallucinations: plausible but incorrect outputs.<br>• No real-time knowledge connection.<br>• Lacked agency: cannot act or make decisions on its own. | • **Upgrade:** LLMs with the ability to call external tools/ services.<br>• **Mechanism:** LLM + API frameworks (LangChain, LangGraph, OpenAI plugins).<br>• **Impact:** API access helps autonomy, real-time knowledge access, and independent decision-making. |
| AI Agent [23,24] | 2020s to present | Tool augmented autonomy | LLM + API Integration | • Limited context awareness; agent myopia.<br>• Task specific; narrow domain knowledge.<br>• Difficulty in choosing the correct tools/APIs and Coordination complexity. | • **Upgrade:** Collaborative multi-agent systems.<br>• **Mechanism:** Shared memory buses, agent orchestration, dynamic task allocation.<br>• **Impact:** Rich context, division of labor, negotiation, and cooperative problem solving. |
| Agentic AI [25] | Emerging | Cognition as service, goal-directed autonomy | Multi-agent orchestration | • Open questions on ethics and safety.<br>• Inter-agent communication costs and conflicting goals. | N/A |

### *3.1 Definition of Agentic AI and Key Aspects*

Agentic AI is an autonomous artificial entity capable of perceiving multimodal input from a dynamic environment, generating strategic objectives through intelligent reasoning and planning, pursuing objectives by taking action independently, and adapting continuously through experiential learning while coordinating with multiple agents inside the agentic ecosystem. This denotes a distributed intelligence and a goal-driven, adaptive ecosystem of autonomous agents, governed by a central orchestrator, that solve complex problems independently with minimal human intervention. The five key aspects defining its operational and distinguishable characteristics include:

- *Autonomy*: Refers to the ability to operate independently, make strategies, select resources, and take actions or decisions without needing constant human involvement. The system can continuously assess its performance, detect errors or handle exceptions, and autonomously solve the issues [27].
- *Goal-oriented behavior*: Reflects the capability of the system to generate clear goals to achieve desired outcomes. The system breaks down complex tasks into smaller steps or goals and optimizes actions or strategies, ensuring consistent focus towards successful task completion [5].
- *Perception and interaction*: Indicates the system's ability to continuously sense and respond to its surroundings effectively. This involves the perception of multimodal data from the environment, proper interpretation of that data, and production of necessary commands to take physical actions accordingly [7].
- *Multi-agent coordination*: Highlights the system's capability to coordinate multiple agents to work together to achieve a shared goal. This requires collaboration among multiple specialized agents through structured communication to share data, resolve internal conflicts, and make decentralized decisions, thereby forming a distributed intelligence network [28].
- *Adaptability and learning*: Demonstrates continuous system improvement through accumulated experiences gathered from previous tasks or human feedback to adapt to new data or shifting conditions. Through mechanisms like feedback loops, meta-adaptation, or advanced learning mechanisms, an Agentic AI system can evolve over time to stay practical and relevant in dynamic conditions [29].

### *3.2 Expanded Comparison among AI Paradigms*

A systematic comparison between traditional AI, Generative AI, AI agents, and Agentic AI across multiple dimensions, including foundational capabilities & autonomy, knowledge & architectural components, operational & interactions, performance & deployment, and risks & limitations, is presented in Tables 2 to 6. The comparative analysis, in which everything before Generative AI is considered traditional AI, revealed that traditional AI is effective only on static, rule-based tasks where the system is not exposed to significant input variance and only predefined outputs are required. However, traditional AI lacks flexibility and is hindered by poor generalization, limited scope, and limited adaptability.

On the other hand, Generative AI is superior at one shot creation-taking prompt and producing an output, but it stops there. It has no memory of the past, cannot interact with tools or external environments, cannot evaluate its own output, cannot self-correct, and cannot break a complex goal into multiple steps. These three pain points in the real world are completely unaddressed: 1) long-horizon planning, 2) reliable execution in a dynamic environment, and 3) minimal human intervention. By contrast, Agentic AI solves these issues by closed-loop architecture: perceive → plan → act → observe → reflect → repeat. Such meta-cognition, self-correction, and reliable execution of complex tasks bridge the gap between producing information and taking verifiable action.

**Table 2:** Foundational Capabilities & Autonomy

| Dimension | Traditional AI | Generative AI | AI Agent | Agentic AI |
|---|---|---|---|---|
| Perception | Structured Data Pattern | Multi-modal input prompts (text/images) | API/Sensor data integration | Holistic Environment sensing |
| Reasoning | Rule-based inference | Probabilistic: Implicit pattern matching in latent space | Task-specific: Decision logic or Single-step LLM prompts | Complex: LLM-backed multi-step reasoning and agent debate |
| Planning | None: Predefined pipelines | None: One-shot Generation | Basic: Single-task sequencing | Advanced: Multi-step goal decomposition, sequencing actions, and dynamic re-planning |
| Autonomy | Low: Need human intervention | Low: Only reacts to the prompt, but no self-initiation | Moderate: Independently carries out tasks when triggered | High: Proactively initiates, monitors, and pursues goals |
| Decision scope | Fixed | Context-limited | Task-bound | Broad and dynamic |

AI agents, on the other hand, can execute tasks on their own; when a multi-domain problem arises, it lacks the lateral awareness to delegate. Furthermore, a single AI agent can hallucinate, leading to a system crash or an infinite loop. Even when multi-agent systems are deployed without an agentic system, they cannot share intermediate reasoning, leading to redundant processing and conflicting actions. This makes AI agents insufficient for enterprise-grade, complex task resolution compared to Agentic AI in addressing real-world problems.

### *3.3 Distinguishing Agentic AI from Closely Related Fields*

To establish conceptual clarity, agentic AI needs to be clearly differentiated from closely related paradigms, such as multi-agent systems, autonomous AI systems, and LLM-based agents. These paradigms share similar goal-directed autonomy and interaction, yet they differ in scope, architecture, and implementation strategy. For instance, traditional multi-agent systems are based on pre-programmed rules, deterministic heuristics, and strict communication protocols; however, Agentic AI does not depend on hand-coded rules for reasoning and dynamically adapts to open-domain challenges. Secondly, autonomous AI systems excel at continuous execution and are confined to narrow, bounded environments. In contrast, Agentic AI operates across unbounded digital and analog ecosystems, capable of abstract and cross-domain planning. Finally, LLM-based agents are reactive, prompt-driven tool executors that suffer from context degradation and require constant human correction. Conversely, Agentic AI is defined by its proactive meta-cognition, utilizing centralized orchestration, persistent layered memory, continuous self-reflection, autonomous task decomposition, execution, and self-correction, as well as long-horizon objectives that are entirely independent of human prompting.

**Table 3:** Architecture & Knowledge

| Aspect | Traditional AI | Generative AI | AI Agent | Agentic AI |
|---|---|---|---|---|
| System Structure | Monolithic Pipeline | Single model API | Model + Tool modules | Orchestrator + Agent Network |
| Memory system | None | Context window | Session-based | Shared knowledge graph |
| Knowledge base | Static, relational DBs | Pre-trained, encoded in a model-based weights | Task-specific DBs | Shared vector DB + knowledge graph |
| Context window | Feature vector | Tokens and prompt history | Session state (short term) | Long-term semantic memory |
| Tool integration | None | Plugins (optional) | Task-specific APIs | Extensive and dynamic tools federation |

**Table 4:** Interaction & Operational Characteristics

| Aspect | Traditional AI | Generative AI | AI Agent | Agentic AI |
|---|---|---|---|---|
| Trigger Mechanism | Data Input | User prompt | Event or command | Strategic goal |
| Workflow | Linear pipeline | Single-step generate | Linear task chain | Multi-agent, branching workflow |
| Error Handling | Fail-silent | Hallucination mitigation | Basic retries | Self-correction protocol |
| Validation Mechanism | Accuracy metrics | Human review | Result verification | Cross-agent or multi-layer critics, human in the loop |
| Human in the loop | For each run | Review outputs | Needed for the new task | Only for exceptions |
| Feedback Integration | Limited, Manual | User-driven | Task-based | Real-time, Adaptive |

## 4 Architecture and Workflow of Agentic AI

Apart from monolithic models, the architectural workflow of Agentic AI represents a cognitively distributed framework with specialized multi-agent coordination, large language models (LLMs), and adaptive learning mechanisms to achieve actionable autonomous intelligent behavior. Figure 2 illustrates the architectural framework of Agentic AI, and Figure 3 depicts the layered design from user interaction to system evolution via processing multi-modal inputs, semantic analysis by LLMs, a structured knowledge base, centralized coordination, distributed multi-agent execution, adaptive response synthesis, and continuous monitoring and governance. The following sections will describe each architectural component, inter-dependencies, and roles in creating a self-optimizing ecosystem that extends beyond just mirroring human-like autonomy.

**Table 5:** Performance & Deployment

| Aspect | Traditional AI | Generative AI | AI Agent | Agentic AI |
| --- | --- | --- | --- | --- |
| Latency | Low (ms-s) | Low-Moderate (ms-s) | Moderate (s-tens of s) | Variable (s-min) depending on workflow complexity |
| Reliability | Predictable, stable | Depends on the prompt quality | Depends on the tool/api availability | Depends on all agents |
| Scalability | Low: scales by retraining models | Moderate: Scales with bigger contexts and computes | Moderate: Scales with APIs or task complexity | High: Scales through distributed multi-agent and load balancing |
| Development complexity | Moderate- ML pipelining | High- Model training and tuning | High- Model + integration logic | Very high- multiple agents, orchestration logic |
| Infrastructure Needs | CPUs | GPUs | Cloud APIs | AI-optimized clusters |
| Deployment model | Centralized, Static | Cloud-centric, scalable | Hybrid, event-driven | Distributed, orchestrated |

**Table 6:** Limitations & Risks (L & R)

| L & R Aspect | Traditional AI | Generative AI | AI Agent | Agentic AI |
| --- | --- | --- | --- | --- |
| Key limitations | Rigid boundaries | Hallucinations, copyright infringement | Narrow capability | Complexity overhead |
| Ethical Risks | Bias in training data | Infringement | Unintended actions | Uncontrolled goal pursuit |
| Security | Data poisoning | Prompt leaking | API breaches | System takeover |
| Failure Modes | Out-of-scope fail-stop, recovery | Factual inaccuracy, limited control over outputs | Tool execution errors | Cascading multi-agent failures |
| Safety and control | Manual recovery | Limited control over outputs | Needs safeguards per task | Requires robust governance, kill-switches, monitoring |
| Explainability | Often transparent | Opaque deep models | Mixed transparency | Difficult to trace multi-agent locks |

### *4.1 System Overview*

An Agentic AI system is structured as a multi-layered framework that interfaces with users and external systems through an adaptive gateway that processes multi-modal inputs ranging from natural language queries to structured API calls and culminates in practical insights, actionable outputs, or results. At its core, LLMs are the brain that drives the understanding, reasoning, planning, and action upon diverse input. At the same time, the orchestrator oversees the task assignment and coordinates the activities of individual agents. The information or context needed for planning or decision-making is supported by the system's knowledge base, which includes long-term memory, short-term memory, and task-specific information.

#### *4.1.1 User/System Interface*

The user/system interface acts as a bridge between external input and results. This interface is designed to handle a wide range of modalities—text, speech, vision, and sensor streams—and translate them into a standardized format to ensure compatibility with the system's perception layer. The interface design ensures fluidness, transparency, and user-centric appearance, prioritizing ease of use and clarity. The symmetric and asymmetric communication between input and output allows users to submit queries or needs and get responses in real-time through fast or delayed processing based on the task. The internal real-time processing mechanism of input and step-by-step synthesis of output can be presented as a summary to the user to ensure a transparent ecosystem. The additional feedback integration mechanism helps tailor tasks to users' specified preferences, constraints, and priorities. This layer also prevents unwanted system access through initial authorization or verification by permitting only intended users and denying access to unauthorized entities

#### *4.1.2 Core Cognitive Engine: LLMs*

The meta-cognitive capabilities of LLMs are the core engine behind the transformation of multi-modal inputs into meaningful, executable intelligence. These models are trained on vast corpora of datasets. They not only process language but also anchor the cognitive backbone of the entire system. Starting from the interpretation of the complex multi-modal inputs, they can perform multi-step reasoning on problems, integrating knowledge from diverse sources, and are capable of generating need-based dynamic responses, which makes them indispensable to the system's operation.

Semantic Understanding and Abstraction:

LLMs represent words, phrases, images, or other data types in a continuous vector space where semantic similarities are preserved through attention mechanisms [22] and the use of embeddings [30]. This empowers LLMs to discern user intent and grasp contextual nuances. As a result, they can interpret complex user queries, extract key insights from knowledge sources, and generate high-level abstract goals by capturing the relationship between the queries and hidden patterns of raw data.

Multi-Hop Reasoning and Planning

Beyond syntactic parsing, LLMs in Agentic AI are tasked to convert the abstract goals, generated from the earlier layer, into cognitive workflows through a sequence of action planning and multi-hop reasoning. Leveraging methods like Chain-of-Thought prompting [31] and self-reflective decoding, LLMs simulate human-like step-by-step reasoning. They also evaluate the consequences of each hypothetical action, use previous experience, and generate or adjust plans accordingly. This allows the system to handle multifaceted tasks with strategic thinking.

Personalization and Memory Recall:

Similar to human memory that continuously gathers and updates existing knowledge by new experiences, LLMs' interaction evolves with the user profile's past interactions, preferences, history, and behavioral cues that are stored in memories. The relevant knowledge from external repositories, structured databases, and ontologies can also be pulled by retrieval-augmented [32] mechanisms to maintain continuity across sessions and to generate more natural and relevant responses that evolve with personalized context.

Knowledge Integration and Justification:

Modern LLMs integrate external knowledge from diverse sources with internal systems through accessing external databases, APIs, or knowledge graphs. The integration process involves evaluation of information sources and cross-referencing, which allows the model to remain up to date on genuine real-time information. They also provide transparency by showing how the conclusion was reached through offering logical pathways or citing state-of-the-art studies.

*4.1.3 Multilayered Knowledge System (L, S, TS)*

Rather than using monolithic architecture, the knowledge system of Agentic AI is organized into multiple layers, based on information of different lifespans, operational relevance, and contextual specificity. This architecture mirrors the human memory hierarchy, which enables efficient storage usage and information retrieval[33].

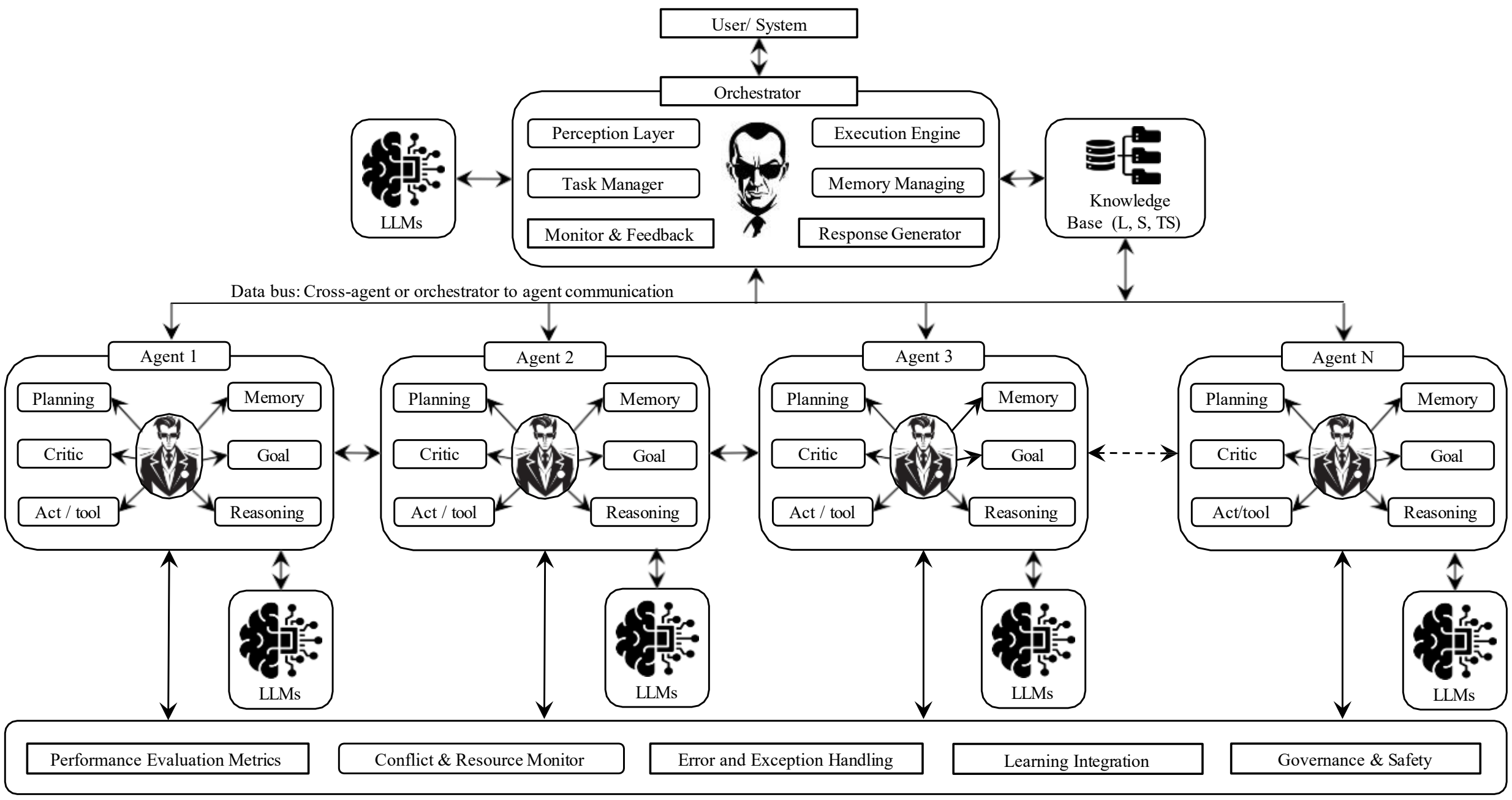

**Figure 2:** Architectural framework of Agentic AI.

Long-Term Knowledge (L):

The long-term knowledge serves as the persistent cognitive memory of an Agentic AI system that consists of fixed or unchangeable information such as learned patterns, rules, ontologies, protocol definitions, scientific principles, historical performance data, and established policies. These are stored in persistent graph-based knowledge stores (e.g., RDF/OWL databases) and optimized for semantic querying and inference [33]. Updates to this layer are infrequent and usually happen via curated batch processes. This memory acts as the foundational cortex and allows the system to leverage decades of aggregated facts or data.

Short-Term Knowledge (S):

Short-term knowledge or session-oriented dynamic memories are used to capture the ephemeral information of ongoing interactions or current sessions. This dynamic memory includes contextual embeddings, recent user queries, temporary inferences, tool usage records, and execution stack traces. Methods like Redis, in-graph caches with low-latency indexing, are used to implement this memory so that real-time, rapid adaptation to user intentions can happen without polluting the long-term memory.

Task-Specific Knowledge (TS):

On the other hand, mission-specific operational knowledge or task-specific knowledge holds structured goal trees, execution flows, runtime variables, agent delegation maps, tool interface specifications, and intermediate outputs. This memory state is highly dynamic and is subjected to frequent reading/writing as the task progresses. This layer enables agents to complete their sub-tasks accurately, monitor runtime, audit performance, and send relevant summaries to long-term memory for future learning upon task completion.

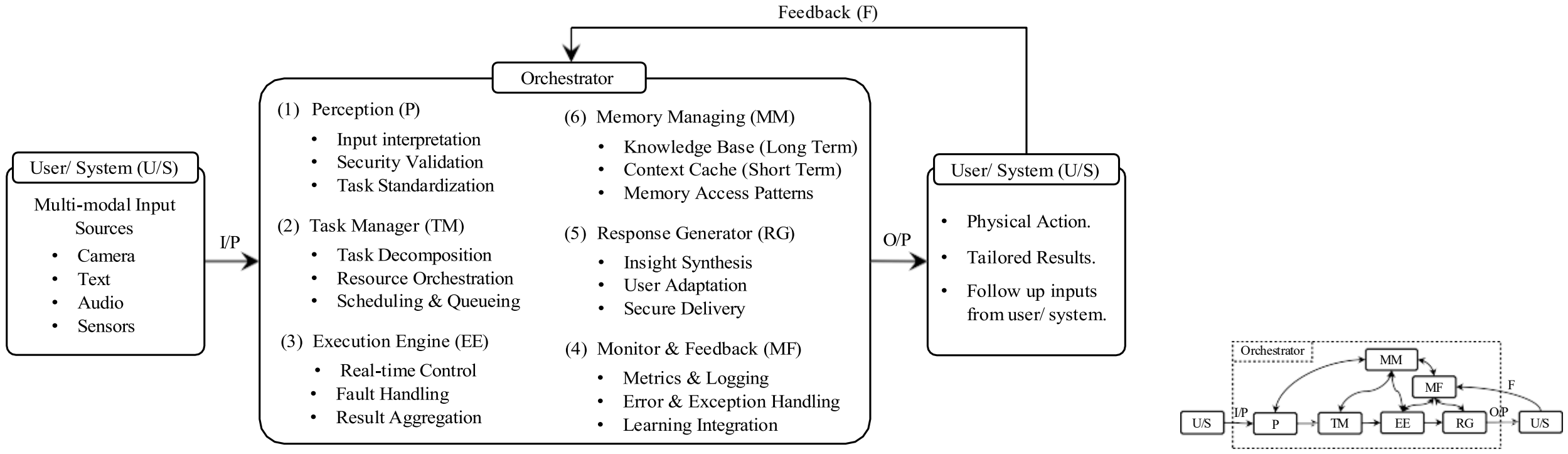

**a** Interaction between the user/system and the orchestrator **b** data flow of each layer

**Figure 3:** Overview of the Agentic AI system's sequential data-flow

### *4.2 Orchestrator: Central Coordination Layer*

The orchestrator serves as the central nervous system of the architecture, leveraging the six-stage workflow to interpret user/system directives, decompose complex tasks, monitor progress, and synthesize tailored responses, all while maintaining a layered knowledge base and continuous adaptation through feedback as described in Figure 2. This integrates neuro-computational principles with industrial-grade reliability to achieve faster decision cycles than disconnected architectures while maintaining low latency for critical operations[34–36].

#### *4.2.1 Perception Layer*

This layer serves as the initial cognitive gateway that leverages LLMs or multi-modal fusion algorithms to interpret and contextualize raw multi-modal inputs into a structured representation suitable for subsequent processing, enabling intelligent task initiation.

Input Interpretation:

Inputs like cameras, text, audio, and sensor streams are converted into unified semantic graphs that encapsulate the user's intent with high fidelity. For instance, textual commands are normalized and tokenized using a sub-word algorithm (e.g., Byte-Pair Encoding) [34]; lemmatization and custom lexicons ensure that idiomatic expressions, code-mixed phrases, and domain-specific jargon are accurately recognized. Transformer-based speech-to-text pipelines remove the background noise, detect the language, and produce a clean transcript for speech commands. Visual information extraction, such as object recognition or reading text via optical character recognition, is done through convolutional or transformer architectures [22]. Time series data or sensor readings are handled via 3D convolution, temporal attentions [35], or timestamp alignment to enable unified reasoning across different modalities. To enable unified reasoning across different modalities, a multi-modal encoder can be used to embed features from text, audio, and vision into a shared latent space [36].

Security Validation:

The system may receive illogical, invalid, or malicious inputs, which will increase computational cost, might crash the system, and generate output that might violate ethics or privacy. That's why security validation is a critical function that employs zero-trust architectures [37] with real-time threat or anomaly detection [38], verification, access control, and policy checking to authenticate input sources. This feature will filter out potential threats before forwarding inputs to the task manager, fortifying the system's resilience.

Task Standardization:

Different users/systems might phrase the same request or input in numerous ways. Instead of addressing them individually, it's more efficient to transform user intent into a standardized task format. This involves mapping raw intent to predefined task schemas and enriching them with contextual metadata drawn from the memory manager. As a result, downstream modules don't have to deal with dozens of input formats; they follow the schema.

#### *4.2.2 Task Manager*

Clean, standardized data from the perception layer is passed down to this layer. The task manager is a strategic planner who distributes complex objectives to appropriate agents while optimizing resource allocation across the agent ecosystem.

Task Decomposition:

The complex user intents or directives are converted into granular, well-defined jobs or sub-tasks using hierarchical planning algorithms or building a directed acyclic graph (DAG). Here, dependency analysis between input and expected output, constraint satisfaction techniques, and fallback strategies are employed in the partitioning task. The granularity of sub-tasks is dynamically adjusted based on agent capabilities and computational load, which enhances the overall task completion efficiency[39,40].

Resource Orchestration:

Each agent or a group of agents of the agent ecosystem possesses a unique skill set. There's no certainty that a particular skilled agent is available then, so the system has to check how busy the targeted agents are and assign tasks intelligently to prevent agent bottlenecks. Considering these, this sub-component allocates sub-tasks to agents based on their skill set or capability profile, current load, and service-level agreements [39]. This sub-component also ensures fairness in resource usage so that no single agent can hog all the available resources.

Scheduling & Queueing:

Often, particular sub-tasks must be executed before other sub-tasks and vice versa. This temporal execution of sub-tasks is optimized and scheduled based on their importance, relevance, and urgency by employing priority-based queueing, critical path elevation [40], dependency-aware deadlock prevention [41], and a dynamic rescheduling algorithm [42]. By applying the start and finish times of each sub-task, this component enables the system to complete jobs before the deadline and reduce latency. On the other hand, if a task fails or a new dependency arrives, this component reroutes the sub-tasks on the fly, so nothing gets stuck, and the system can adapt quickly to new changes or errors.

#### *4.2.3 Execution Engine*

This component is the operational core engineered to coordinate agent activities, task execution, maintain a fault-tolerant distributed cognition system, and is responsible for aggregating the output of all the agents into a single, reliable result through precision control mechanisms.

Real-time Control:

This sub-component supervises agent execution and helps synchronize agent actions through a control loop. This ensures that interdependent sub-tasks are executed in lockstep and can minimize execution delays by dynamically allocating more computational resources where needed to accelerate the process by reducing resources in other sub-tasks, allowing the engine to perform faster and maintain temporal coherence across distributed agents.

Fault Handling:

To prevent the system from crashing, the execution engine needs redundancy or a recovery plan to handle agent failures or network disruptions. If an agent fails, automatic retries from agent-level checkpoint rollback [43] can help, as they address transient glitches. If not, backup agents must take over, as a spare car tire, or workflow reconfiguration will occur automatically. The system will raise alerts for persistent issues so a human engineer can step in and fix them.

Result Aggregation:

After completing each sub-task, agents pass the output to the orchestrator. If two agents return slightly different outputs, then confidence-weighted consensus is used. If distributed outputs are returned from agents, employing data fusion techniques [44], the orchestrator ensures all the pieces are stitched together. If opposite or conflicting answers arrive, this sub-component statistically reconciles and provides coherent answers to the next module.

#### *4.2.4 Response Generator*

The presentation layer of an agentic system is the response generator that synthesizes and delivers the final output by transforming aggregated results into contextually appropriate, secure, and user-friendly responses[45].

Insight Synthesis:

Aggregated results are converted into meaningful, clear deliverables such as written summaries, simple charts, or actionable commands—so users or downstream systems can immediately understand and act on them.

User Adaptation:

This sub-component delivers responses that match each user's expectations, making them more engaging and easier to act on. The user embedding vectors are stored and retrieved from a vector store [45] that contains user preferences, profile, and interaction history. Personalized responses are generated by adjusting LLMs' tone based on the data.

#### *4.2.5 Monitor & Feedback Mechanism*

To enhance future operations, the meta-cognitive oversight system enables self-calibration through adaptive refinement, system performance evaluation, and learning integration[46,47].

Metrics & Logging

Performance indicators such as task completion times, resource utilization, error rates, knowledge recall accuracy, cross-agent transactions, and agent-level performance indicators, along with distributed logging frameworks that record agent activities in real time, enable detailed post-hoc analysis. Advanced analytics algorithms process these metrics to identify trends, anomalies, and performance-degradation patterns, thereby assessing system health and identifying optimization opportunities.

Self-Healing Triggers

This subsystem enables automated recovery procedures through an intelligent trigger mechanism when performance thresholds are exceeded, or error rates increase beyond acceptable levels. Advanced machine learning models analyze historical patterns to predict potential failures [46]. Once failure is sensed, automated healing procedures, including graceful degradation [47], load balancing [48], resource re-allocations, and workflow reconfiguration, are activated.

Learning Integration

This component continuously integrates operational telemetry (e.g., performance logs, anomaly flags), transfer learning [49], structured user feedback (ratings, comments), and insights discovered during operation to improve future decision-making and system capabilities. Reinforcement learning or online learning [50] algorithms are employed to update the knowledge graphs. The system maintains version control for learned models and can roll back to previous versions if performance degradation is detected.

#### *4.2.6 Memory Manager*

This component functions as the orchestrator's cognitive storage system, which interfaces with other modules such as perception, task manager, execution, and response through monitoring and feedback, so that intelligent information retention, retrieval, and knowledge synthesis can happen seamlessly across the entire orchestrator workflow. Employing neuromorphic memory [51] principles with a distributed storage architecture, the memory manager maintains contextual awareness, preserves institutional knowledge, and supports system evolution.

Short-Term Memory

The orchestrator's active working memory contains conversational context embeddings, current task execution graphs with dependency, agent's metrics and logging, intermediate processing results, feedback, and user preferences. Without this sub-component, the orchestrator would lose track of task dependencies, context, and results, leading to the system's crashing. This memory provides fast access to fresh data, synchronizes between sub-tasks, enables the system to run faster, and automatically performs memory consolidation to transfer important short-term information to long-term memory.

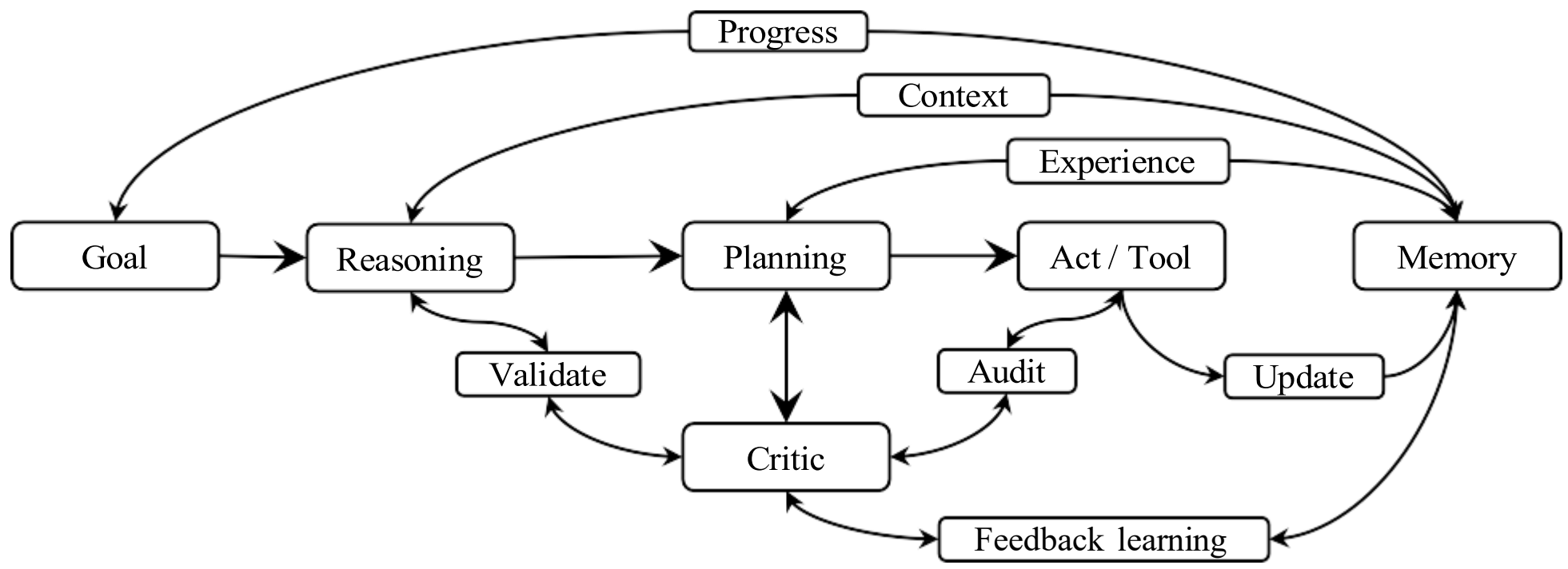

**Figure 4:** Agents' internal cognitive architecture with proper data flow.

Long-Term Memory

This is the orchestrator's persistent knowledge repository, which contains users' interaction patterns, preferences; system's task completion patterns, resource utilization and optimization outcomes; agent's capability with performance metrics, speed, and reliability assessments; successful task decomposition templates with patterns and resource requirements; domain-specific knowledge graphs [52] with entity relationships, concept hierarchies, and semantic associations. Long-term memory transforms a reactive system into an adaptive system that learns and evolves. Without this memory, the orchestrator would repeatedly make the same decisions and mistakes, fail to personalize user experiences, and lose valuable insights gained from system operation.

Memory Access Patterns

To leave sufficient processing time for complex reasoning and task coordination, it's necessary to complete memory access within a microsecond timeframe. Suboptimal memory access patterns would create bottlenecks, increase latency, reduce user satisfaction, and limit the orchestrator's ability in complex reasoning and coordination. That's why intelligent memory access is required to meet real-time demands, balancing speed and accuracy. The memory manager analyzes historical access patterns or uses trends to predict future needs, enables proactive data loading, employs multi-level caching strategies, and uses advanced search algorithms for smooth task handovers.

### *4.3 Multi-Agent Layer*

#### *4.3.1 Agents' Internal Structure*

The cognitive framework of individual agents is presented in Figure 4, which establishes a cyclical workflow for autonomous problem-solving and decision-making. The architecture demonstrated how agent-based AI can achieve goal-oriented behavior through the interconnection of multiple sub-components, such as goal setting, reasoning, planning, action/ tool, critic, and memory[53]. Each component is outlined as follows:

- **Goal Setting:** Agents receive a formalized goal from the orchestrator, which is encoded as tuples $G = (\phi, \Gamma, \Omega)$. Here $\phi$ denotes the objectives to be fulfilled, $\Gamma$ defines temporal, resource, or domain-specific constraints, and $\omega$ represents the logical or quantitative success thresholds. This explicit goal abstraction ensures objectives remain verifiable and aligned with system-wide intention. This component maintains a direct connection with memory to monitor goal-achieving progress and further goal refinement or adaptation through the stored feedback from the user/system.
- **Reasoning:** The reasoning module maintains a direct connection with the goal component from where it receives primary input to trigger the cognitive process. The reasoning module also has a bidirectional connection with memory for a unified context graph through which agents leverage historical knowledge, past experiences, and learned patterns. This procedure ensures accumulated wisdom-based reasoning rather than isolated operation. The contextual synthesis is converged with probabilistic hypothesis validation (confidence score) for quality assurance of the reasoning itself. In this way, logical consistency is maintained, and flawless decisions are finalized.
- **Planning:** The planning module converts the finalized insights from the reasoning module into concrete, executable action sequences. This module employs algorithms such as A* search or Monte Carlo [53] tree search to formulate a sequence of step-by-step action plans to achieve the established goal. The planning phase is not solitary; rather, it incorporates past experience from memory, which allows the agent to learn from previous successful and failed plans. The bidirectional connection with the critic module enables continuous plan evaluation and refinement.
- **Acting/Tool Invocation:** This component converts strategic plans into concrete, tangible action and serves as the primary interface between the Agentic AI system and the external world. The act/ tool component receives primary inputs from the planning module and executes the step-by-step action sequences by calling external tools, APIs, or controlling the environment. During execution, this component generates progress information of action and effect, which are stored in memory for operational awareness. Additionally, the auditing mechanism ensures quality control, ensuring that actions are performed correctly and efficiently.
- **Critic:** This layer serves as the reflective and evaluative conscience framework, which continuously refines the agent's cognitive processes. All the major components, such as Reasoning, planning, act/tool, and memory, are connected with the critic module. It assesses decisions of the reasoning module, reflects the planned workflow of the planning module, evaluates the actual outcomes against the expected results of the act/tool module, and identifies areas where improvements are needed. Finally, a feedback learning loop from critic to memory contributes to agent learning and adaptation through negative rewards for failures and positive rewards for efficiency.
- **Memory:** Agent-specific memory module serves as the dedicated cognitive workspace's persistent repository and learning substrate. This module holds the goal provided by the orchestrator and updates from the act/tool about task progress, which helps determine task completion rate. The contextual information of the memory module provided to the reasoning module helps in data-informed decision-making. This component provides previous successful and failed planning patterns to the planning module to formulate an informed action plan. It is also connected with the critic, and the critic's assessment helps in storing persistent knowledge based on agent experience, which might flow to the long-term memory of an Agentic AI system.

#### *4.3.2 Agent Interaction*

In Agentic AI systems, agents do not operate in isolation; they form a communicative framework to transcend their isolated capabilities and participate in collective intelligence. These interactions involve information exchange, collaborative decision-making, and coordinated action execution, broadly falling into two categories: direct agent-to-agent and agent-orchestrator collaboration.

Cross-Agent Collaboration

This type of collaboration represents direct peer-to-peer communication through shared message buses, direct peer calls, or mediator-based interactions. The collaboration framework implements a negotiation protocol with a capability matching algorithm to pair agents with complementary skills, a resource-sharing agreement to optimize collective resource utilization, and a conflict resolution mechanism to reconcile contradictory objectives or outputs. Distributed consensus formation, collective problem decomposition, and load balancing occur naturally between agents without explicit central planning due to cross-agent collaboration. Effective collaboration can optimize resource sharing, faster task completion, and improved system performance.

Agent-Orchestrator Collaboration

The orchestrator's feedback loop establishes a bidirectional communication channel for information exchange between agents. The orchestrator assigns tasks to the agents, and agents send feedback tuples ($\tau$, $\psi$, $\omega$) upon task completion. Here, $\tau$ indicates task completion status (e.g., success/failure/timeout), $\psi$ indicates the performance metrics (e.g., F1-score, regret, resource consumption), and $\omega$ represents causal traces for explainability and optimization for future task allocation. It empowers the orchestrator to detect inefficiencies—e.g., reallocating tasks from an overloaded agent—and supports learning algorithms (e.g., reinforcement learning) for future strategy refinement. This creates a closed-loop system where individual agents share their experiences with collective intelligence, and collective insights enhance individual agent performance.

#### *4.3.3 Coordination Topologies*

The coordination topologies define the structural, operational, and communication patterns that govern the interaction of agents within the cognitive system. A visual representation of six different topologies is depicted in Figure 5. These topologies represent the fundamental architecture for agent connections, information flow, and decision-making processes that influence collective influence, system scalability, resilience, and efficiency.

Hierarchical Orchestration (Tree Topology)

This is a centralized coordination model also known as tree topology, where a single orchestrator acts as the tree's root and serves as the supreme coordinator. Below the orchestrator, agents are organized into layers, with higher-level agents (e.g., manager agents) delegating tasks and overseeing a specific domain or group of worker agents.

**How it works:**

- Task initiation: The orchestrator generates the goal as per user/ system demands.
- Task Decomposition: Orchestrator further decomposes the goal into sub-tasks and assigns them to manager agents.

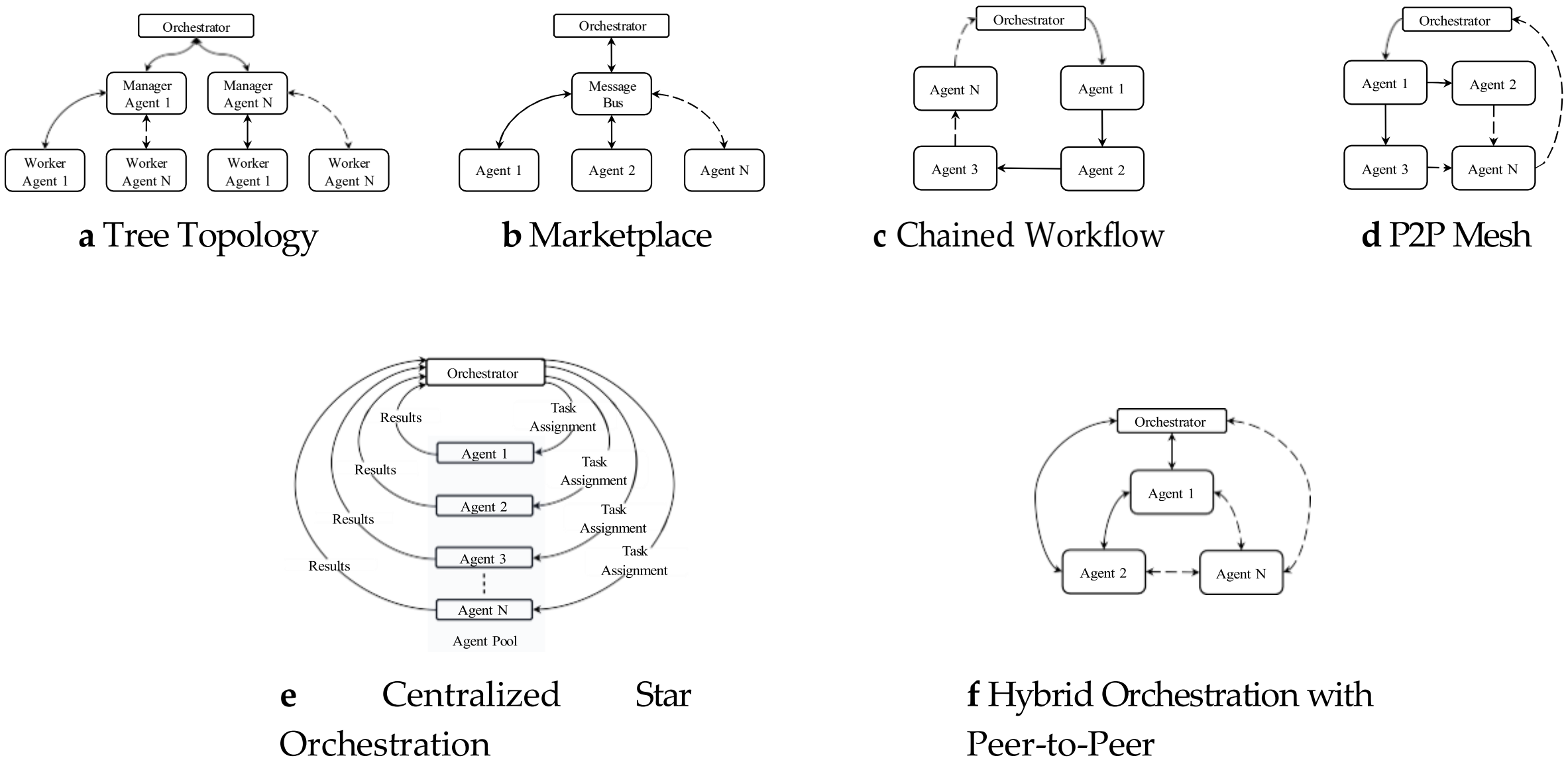

**Figure 5:** All the possible orchestrator-agent coordinating topologies of agentic AI.

- Delegation: Manager agents further delegate sub-tasks to their subordinate agents (e.g., workers).
- Execution and feedback: The subordinate agents process their tasks and report results to the manager, who then reports back to the orchestrator.

**Merit:**

- Clear structure: The hierarchy provides a clear chain of command, which simplifies task management and accountability.
- Centralized oversight: The orchestrator maintains a global view, facilitating monitoring and decision-making.
- Scalability within layers: New agents can be added at the lower levels without overwhelming the orchestrator.

**Demerit:**

- System failure: If the orchestrator fails, then the entire system collapses. If a higher-level agent fails, then the corresponding sub-layer also fails.
- Potential Bottleneck: High task volumes or communication demands can overload the orchestrator.

**Use Cases:** Military command structures where centralized control ensures orderly execution, corporate hierarchies where tasks are delegated to teams in an organized workflow, and a centralized computing system.

Decentralized Coordination (Marketplace)

This is a distributed model in which agents interact directly through a marketplace (message bus) to negotiate or bid for tasks and resources. This topology offers a decentralized architecture in which coordination emerges from individual agent decisions rather than top-down planning.

**How it works:**

- Task announcement: Tasks are posted by the orchestrator to a shared platform (e.g., message bus).

- Agent negotiation: Agent assesses the tasks based on their capabilities, market conditions, cost, demand patterns, and negotiates directly with peers for task bidding.
- Task execution: The assigned agent completes the task without explicit centralized control.
- Result sharing: Outcomes are shared among agents or delivered to the orchestrator through the marketplace.

**Merit:**

- Fault tolerance: If one agent fails, the other agent might take over. So, loss of an individual agent does not halt the system.
- Scalability: Adding new agents to increase system capacity is easy and requires little energy.
- Adaptability: Competitive pressure encourages innovation of new capabilities and improved service quality.

**Demerit:**

- Negotiation delay: Time spent on negotiation can delay task completion.
- Coordination complexity: The need for a sophisticated negotiation protocol increases operational overhead.
- Suboptimal outcomes: Lack of central oversight may lead to inefficient task allocation.

**Use Cases:** Freelancing platform where agents bid for jobs directly, or a cloud computing platform that uses a marketplace for allocating computing resources.

Chained Workflow

This is a linear coordination model where tasks are processed sequentially from Agent1→Agent2→...→Agent N. Each agent performs a specific operation, contributing expertise to the collective problem-solving effort, which allows high-quality outcomes through the systematic application of focused knowledge and skills.

**How it works:**

- Workflow definition: Orchestrator provides information to form the agents in a cyclic order and assigns a task to the first agent.
- Sequential execution: The first agent processes the triggering input from the orchestrator, passes the output to the next agent, and continues until the last agent.
- Output delivery: The last agent delivers the final result.

**Merit:**

- Simplicity: The design is straightforward, easy, and fault identification is simple.
- Linear efficiency: Each agent can focus on its expertise and rely on other agents for complementary capabilities, allowing a diverse expert ecosystem while maintaining consistent processing quality.
- Traceability: Progress can be easily monitored, and bottlenecks can be easily identified.

**Demerit:**

- Inflexibility: A slight change requires redesigning the chained architecture.
- No parallelism: Sequential execution does not allow parallel task processing, so heavy task completion will be slowed down.
- System failure: If one agent of the agent loop fails, the overall system collapses.

**Use Cases:** Data processing pipeline, software development, and manufacturing assembly, where sequential workflows are needed for task completion.

Swarm Intelligence (P2P Mesh)

This decentralized model, also known as peer-to-peer (P2P) mesh topology, is inspired by systems like ant colonies. Here, agents autonomously interact with neighbors to share discoveries or state signals without a central coordinator. The orchestrator only intervenes for initial goal injection, global configuration, and result aggregation.

**How it works:**

- Rules establishment: All agents follow simple rules related to task assignment and agent interaction.
- Peer-to-peer interaction: Direct communication with neighboring agents forms a hybrid network, where each agent communicates with every neighbor that they seem fit.
- Emergent behavior: Collective action-based results emerge from previous agents to the next agent to solve a global problem. The system adapts dynamically as agents adjust their behavior based on their experience with task completion. The final output is sent back to the orchestrator.

**Merit:**

- Resilience: No single agent failure will collapse the system.
- Scalability: Effective with a large agent population, as coordination is local.
- High adaptability: Thrives in a dynamic or uncertain environment due to self-organization.

**Demerit:**

- Unpredictability: Emergent behaviors and resource requirements can be difficult to predict, and fault identification is hard.
- Design complexity: Crafting rules for agent interaction and dynamic data traffic coordination requires advanced engineering.
- Resource intensity: Large agent colonies will need heavy resources for task completion.

**Use Cases:** Distributed sensor network, autonomous vehicle coordination, and a financial trading system that implements swarm intelligence.

Centralized Star Orchestration

The orchestrator broadcasts the task to specialized agents, each of which contributes to task execution, and sends the results back to the orchestrator. No agent-to-agent communication occurs; all control and feedback traverse the Orchestrator, which serves as the sole nexus.

**How it works:**

- Task assignment: The orchestrator assigns a role to each of the agents and decomposes complex tasks into sub-tasks to assign them to agents.
- Execution and contribution: Each agent completes the task, and the results are sent back to the orchestrator.
- Coordination and synthesis: The orchestrator supervises the agents, aggregates their results, takes the informed decision or action, and integrates feedback for environment adaptation.

**Merit:**

- Simplicity of coordination: Centralized control simplifies management, auditing, and policy enforcement.
- Scalability: Adding or removing agents is simple and does not require altering the entire architecture.
- Collaborative synergy: The interplay of specialized agent-based roles fosters collective problem-solving ability with different points of view.

**Demerit:**

- Single point of bottleneck: The orchestrator I/O channel can be overwhelmed by agents' I/O while handling a complex problem.
- Limited agent collaboration: Absence of direct agent-to-agent communication limits fine-grained, collective problem-solving capability.
- Complexity in role management: Assigning roles and managing individual agents increases computational complexity.

Hybrid Orchestration with Peer-to-Peer

This hybrid topology blends the P2P interaction of agents with centralized orchestration. This creates a sophisticated system that maintains the advantages of distributed and centralized coordination while addressing individual topologies' limitations.

**How it works:**

- Centralized oversights: Orchestrator assigns tasks, monitors agents without micromanaging, and resolves high-level issues like conflict or fault management.
- Peer interaction: Agents collaborate independently for data sharing, task synthesis, and completion.
- Execution: On sub-tasks, agents independently take action on their own.
- Feedback loop: Action-based results are returned to the orchestrator for system monitoring, evaluation, and further adjustment.

**Merit:**

- Enhanced resilience: A highly fault-tolerant and adaptive system based upon an intelligent agent ecosystem and centralized management.
- Balanced flexibility: Agents and orchestrator adaptability enhance fine-grained response in dynamic or unpredictable environments.
- Efficient coordination: Local conflict resolution and advanced resource sharing expedite operations while minimizing resource cost.

**Demerit:**

- Orchestrator bottleneck: The system can still choke under heavy load.
- Complex protocol: Requires managing rules for both P2P and orchestrator-to-agent communication.
- Design challenges: Balancing between autonomy and oversight complicates system architecture.

**Use Cases:** Distributed file systems, online gaming, and collaborative AI systems where the orchestrator coordinates autonomous entities.

### *4.4 System Monitoring and Evolution Layer*

This is the foundational substrate upon which the entire observational mechanism and the adaptive intelligence operate. It sits beneath the orchestrator and agent pool, enabling continuous, real-time telemetry analysis, conflict and resource management, error and exception handling, continuous adaptation through learning, and enforcing governance and safety across the whole system. The system monitoring and evolution layer is outlined in detail as follows [54–61].

#### *4.4.1 Performance Evaluation Metrics*

This represents a multidimensional framework for quantitative and qualitative assessment of the system's efficacy, efficiency, and reliability. This subsystem of Agentic AI transcends simple scalar metrics by implementing a hierarchical evaluation framework on individual agents, agent interactions, and system-wide operations levels to gain fine-grained details and tune individual components for full system optimization [54,55].

Agent-Level Metrics

- Reasoning Latency: Time taken to integrate new context and validate hypothesis.
- Planning Throughput: The number of plans generated per second and the average directed acyclic graphs (DAGs) construction time.
- Execution Success Rate: The rate of valid outputs generated from successful act/tool calling within the time-frame.
- Resource Usage: Agent's CPU, GPU, and memory consumption pattern.
- Memory Cache Hit Ratio: Ratio of successful information retrieval from short-term and long-term memory.
- Learning Convergence: The number of training episodes required to reach a certain performance threshold, measured across different task domains.
- Decision Quality: The decision accuracy and confidence score of agents, measured against the ground truth.
- Resilience to Adaptation: Agent's capability of maintaining performance, measured across varying environmental conditions.
- Novel Task Adaptability: Measures the agent's ability to generalize to new and unseen tasks.

Agent Interaction-Level Metrics

- Inter-agent Latency: The round-trip communication delay between cooperating agents during information exchange.
- Conflict Resolution Efficiency: The frequency of conflict occurrence during task execution and the time taken to reconcile between agents.
- Task handover Success Rate: The rate of successful sub-tasks transfer from one agent to another without retries and rollbacks.
- Consensus Convergence Time: The average number of communication rounds required to reach an agreement on a collaborative decision.
- Information Sharing Efficiency: The ratio of successful relevant information transmitted to total information exchanged.
- Load Balance Efficiency: measures fairness in task distribution across agents to prevent a bottleneck.

System-Wide Metrics

- End-to-end latency: Total time taken by the system from goal ingestion to final response.
- Aggregate throughput: The rate of goal completion per minute by the entire agent pool.
- Resource saturation and scaling events: cluster-level CPU/GPU utilization, counts of auto-scaling events, and load-balancer queue lengths.
- SLA compliance: The percentage of target completion (e.g., 90% of tasks completed under 2 s).
- Availability and error rate: System up-time percentage and ratio of critical failures to successful executions.
- System resilience index: Measures the system's performance under adverse and dynamically changing conditions.

- Cross-component synchronization: Measures cross-component performance correlation and synchronization accuracy between components.
- User satisfaction: Measures user experience through user feedback on task completion satisfaction and system usability.

#### 4.4.2 *Agent Evaluation Benchmarks*

Agentic systems demand an evaluation framework that goes beyond static benchmarks like MMLU or HumanEVal. Unlike traditional metrics, which capture isolated capabilities, agentic evaluation focuses on functional corrections, such as whether the agent can complete the end-to-end task, can it recover from error, or maintain state and observe the environment? There are several agent evaluation benchmarks that can be used to test the agentic systems.

- AgentBench: Liu et al. [54] proposed a multi-dimensional benchmark that incorporates 8 distinct interactive environments of 3 types based on real-world scenarios to evaluate LLMs as agents. Agents were scored on their ability to follow instructions, reason, plan, and perform ground actions. Its unified toolkit enables consistent comparison across agents.
- WebArena: Zhou et al. [55] designed a set of benchmark multi-step web tasks (e.g., e-commerce, content management systems). The tasks are diverse, long-horizon, and formulated to imitate tasks that humans regularly perform on the internet. It requires agents to translate high-level natural language directives into HTML DOM manipulations.
- SWE-bench: Jimenez et al. [56] proposed this benchmark for software engineering agents, consisting of 2,294 software engineering problems curated from real GitHub issues. It requires a system to navigate a large database, understand and coordinate multiple classes, functions, and files, process extremely long contexts, and perform complex reasoning. Upon evaluation, Claude 2 was able to solve 1.96% of the issues.
- GAIA (General AI Assistant): Mialon et al. [57] proposed a set of 466 carefully crafted, unambiguous real-world questions. Solving these questions requires multi-step reasoning, multi-modal understanding, web browsing, tool orchestration, and robust error recovery. The current system (GPT-4) scores around 15% compared to humans' 92% accuracy, positioning GAIA as a milestone benchmark for advanced agentic systems.

#### 4.4.3 *Conflict and Resource Monitor*

In Agentic AI, multiple agents and the orchestrator operate concurrently, sharing the same resources, which is why conflict and resource-monitoring subsystems are necessary for efficient resource allocation and for preventing conflicts before they escalate and threaten the system. Conflicts arise in two forms: resource contention (e.g., multiple agents requesting one LLM simultaneously) and logical inconsistencies (e.g., contradictory outputs from agents). To detect conflict, agent interactions are modeled using graph-based analysis to identify potential conflict [58]. Game-theoretic approaches are used to resolve resource conflict where agents are treated as rational actors competing for limited resources. Prioritizing critical tasks based on urgency or system goals, a strategy called Nash Equilibrium is applied to allocate resources efficiently [59]. For example, if Agent 1 and Agent 2 compete for knowledge base access, then Agent 1 should get permission if its task aligns with high priority, queuing Agent 2.

On the other hand, logical conflicts are resolved by consensus algorithms [60], where the orchestrator aggregates and selects the most reliable agent output based on historical performance data, as tracked by the sub-component monitor and feedback of the orchestrator. Automated negotiation protocols are also used to resolve conflicts; if they fail, the system escalates to mediation algorithms that use neutral arbitration logic to propose solutions. In extreme cases, the system depends on an authoritative resolution mechanism to maintain the system's stability.

The resource monitoring system maintains a detailed resource utilization profile for each agent and task type. Advanced algorithms analyze resource consumption patterns to anticipate demand spikes and preempt bottlenecks. The feedback loop between the monitor and orchestrator enables dynamic adjustments by considering immediate resource needs and predicted future requirements, agent priorities, and system-wide optimization objectives. It ensures fairness, efficiency, and resource allocation stability by reallocating or scaling the agent's capacity under load [59,60].

#### *4.4.4 Error and Exception Handling*

This is the immune system of Agentic AI architecture, designed to maintain system resilience and operational continuity by detecting, logging, and mitigating failures[61–63]. This subsystem used a multi-layered approach that combines error classification, resolution strategies, and logging to achieve adaptive error prevention by monitoring overall system components [64,65]. These are described as follows:

Error classification

- Tier-1 (Agent Level): This type of error occurs within individual agents, which have a limited effect beyond that agent, for example, reasoning failures, execution errors, agent's internal memory corruption, learning degradation, and communication failures.
- Tier-2 (System-Level): This type of error affects multiple agents or system components, which have a broader impact on the overall system, such as resource contention, coordination failure, load balancing failures, network partitions, and cascading failures.
- Tier-3 (Critical): This type of error represents system-threatening failures such as safety violations, security breaches, system integrity failures, resource exhaustion, and governance failures.

Resolution strategies

- Automated resolution: This strategy emphasizes an automated recovery mechanism that can respond without human intervention. Some algorithms used in these strategies are graceful degradation, component isolation, automatic restart procedures, fallback operation modes, etc.
- Escalation procedures: When the automated resolution mechanism becomes insufficient, the system implements some escalation procedures such as multi-level arbitration, human-in-the-loop, emergency protocols, and the integration of external resources.
- Adaptive recovery: To make the system even more resilient, it should be able to deal with novel errors that fall outside the resolution procedures. To make it happen, the system synthesizes multiple resolution approaches to address new errors or performs collaborative resolution that integrates distributed intelligence, experimental approaches, and learning-based adaptation.

Post-mortem analysis

This is the learning and improvement component of the error handling system. The post-mortem process begins with root cause analysis of errors by causal chain reconstruction, multi-factor analysis, pattern recognition, and impact assessment. The insights gained from this analysis are integrated into the system knowledge through knowledge base updates, policy refinement, the development of novel strategies for error prevention, and enhancements to the resolution playbook to improve future error prevention and resolution capabilities.

#### *4.4.5 Learning Integration*

Unlike traditional AI, which learns only during the discrete training phase, this subsystem serves as the living, intelligent evolutionary engine of Agentic AI and is responsible for continuous system adaptation. This component transforms every interaction, outcome, and decision into an ongoing learning process that enhances the capability of both the agents and the orchestrator. Multiple learning paradigms, such as reinforcement learning, meta-reinforcement learning, and RAG, are utilized to create this adaptive intelligence framework.

Reinforcement Learning (RL)

A distributed multi-agent reinforcement learning framework employs advanced algorithms such as distributed proximal policy optimization (PPO) [61], multi-agent actor-critic methods [66], and hierarchical reinforcement learning [67]. The agents learn from the interactions with the environment or with other agents and develop behavioral strategies spanning from immediate tactical decisions to long-term strategic planning. The orchestrator provides a global reward signal aligned with task objectives, while the memory manager stores state-action-reward tuples on task completion success rates, user satisfaction metrics, resource utilization efficiency, and collaboration effectiveness. This information not only improves agent-level efficiency but also focuses on the collective efficiency of the entire system over time.

Meta-Reinforcement Learning

This sub-component extends RL by guiding the system to learn to learn more effectively by optimizing an inner-loop adaptation mechanism, enabling the agent to adapt to novel tasks efficiently with minimal training data. Using algorithms like gradient-based model-agnostic meta-learning (MAML) [68], few-shot learning, or recurrent meta-policy, this sub-component develops generalized strategies for knowledge acquisition from recent executions. This meta-cognitive capability optimizes learning efficiency, reduces retraining time, and helps to maintain proficient system performance in dynamic contexts.

Retrieval-Augmented Generation (RAG)

RAG [32] enhances the decision-making capabilities of agents by incorporating external knowledge sources or experiences into internal system memory. This framework maintains distributed knowledge bases that contain factual information, procedural knowledge, and experimental wisdom accumulated through system operations. The retrieval module of RAG uses semantic or vector search to query document stores, vector indices, or ontological graphs to fetch relevant support facts in real time, thereby eliminating problems like hallucination and ensuring that LLM outputs are contextually grounded and accurate, and making the system capable of complex query handling and system evolution.

#### *4.4.6 Governance & Safety*

This layer serves as the critical safeguard ensuring the system's operations remain aligned with ethical guidelines, safety constraints, and regulatory mandates [69]. The governance and safety module interfaces with orchestrator, agents, and all the shared resources via the data bus, enforcing a distributed yet cohesive regulatory framework [69,70].

Policy Enforcement Engine

This constitutes the front line of defense that systematically scrutinizes every action and data flow against a comprehensive set of governance criteria. Each agent's output-weather reasoning, planning, action, or LLM-generated response is evaluated in real time. The engine leverages action control lists (ACLs) to regulate permissions, the General Data Protection Regulation (GDPR) and the California Consumer Privacy Act (CCPA) for data privacy, and model usage guidelines to prohibit disallowed content (e.g., hate speech, misinformation). If this sub-component is triggered, it responds with appropriate protocol, such as action blocking, escalation to human oversight, or system shutdown in extreme cases.

Safety Monitors

This system operates as run-time detectors and ensures system stability or user safety by detecting and mitigating risks such as unsafe command sequences (e.g., deletion of critical infrastructure), bias amplification in outputs, or adversarial attacks. Risk identification is performed using a hidden Markov model, an anomaly detection algorithm, or a counterfactual fairness testing algorithm [70], and remedies are provided according to the system's protocol.

Rule Update Pipelines

This layer updates the governance rules and safety monitors according to the real-world changing standard for system alignment with external regulations and internal governance. The updates propagated through the orchestrator and agent to the whole Agentic AI system. This sub-component allows the system to evolve with the continuously changing standards of the current world and to remain vigilant to emerging threat patterns.

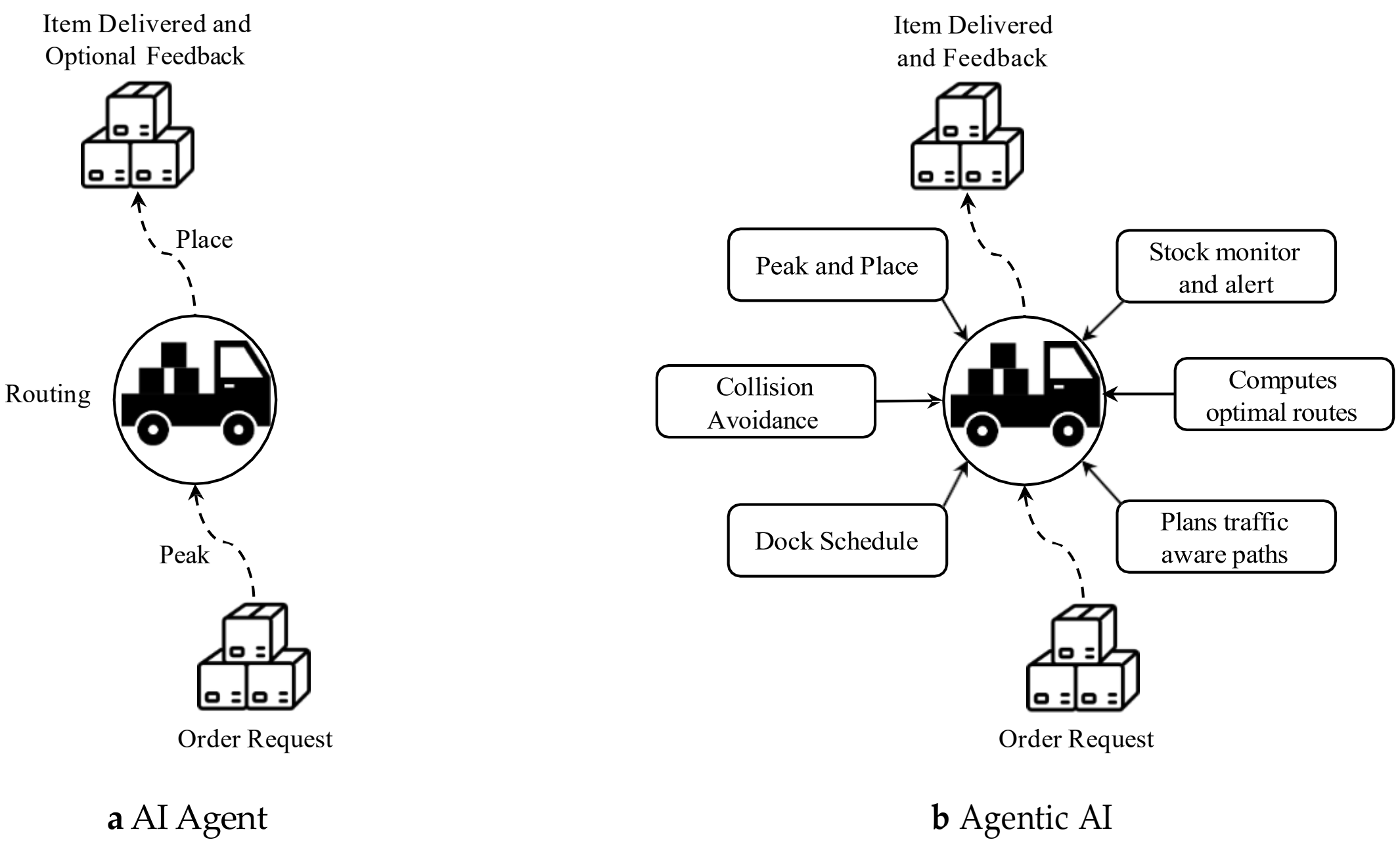

**Figure 6:** Industry-level smart transportation framework using an (a) AI agent and (b) Agentic AI framework, showing the operational and contextual superiority of the agentic system.

### 4.5 Comparative Analysis with the Existing Frameworks

The proposed architecture addresses specific limitations regarding orchestration, memory persistence, and governance of several widely deployed frameworks such as AutoGen, LangGraph, CrewAI, and Semantic Kernel. Autogen largely relies on a conversation-driven paradigm where agents interact through dialogue exchange. While providing flexibility, this scheme suffers from context degradation and a lack of centralized execution in complex deployments. LangGraph operates as a cyclical state machine; though it offers excellent predictability, it rigidifies architecture, which makes dynamic task decomposition cumbersome. CrewAI operates on a role-assigning scheme, which excels at linear task completion but suffers from the absence of a globally accessible knowledge base for agent communication. Lastly, Semantic Kernel primarily integrates LLMs with plugins, which emphasizes traditional software integration over autonomous multi-agent reasoning.

In contrast, the proposed architecture introduces a hybrid neuro-computational approach where a centralized orchestrator isolates six critical operational stages (P, TM, EE, MM, RG, and MF) as shown in Figure 3 for better context awareness, task decomposition, memory sharing, and governance compared to previous frameworks. Unlike AutoGen or CrewAI, which loosely use memory for the agent's context window, the proposed framework utilizes a structured, layered knowledge base (L, S, and TS) controlled by an orchestrator. Furthermore, the inclusion of performance, conflict, resource, learning, error, governance, and safety layers provides industrial-grade reliability that is currently left to manual developers in LangGraph and Semantic Kernel. This centralized cognition over decentralized independent agents ensures robust fault tolerance and system-wide alignment.

### 4.6 Critical BoElenecks of Agentic AI

Despite the theoretical superiority of Agentic AI, the real-world deployment of such a system is currently hindered by a triad of severe bottlenecks: architectural barriers, computational bottlenecks, and ethical and governance deficits.

- Architectural Barrier: The primary barrier is the friction between traditional systems that rely on deterministic APIs, centralized databases, or rigid access protocols with non-deterministic, probabilistic AI agents, which frequently leads to system failures and integration gridlock. Compounding this issue is the struggle of the Agentic system with multi-step reasoning over a long horizon, a fragile tool and agent coordinating pipeline, and low trust in high-stakes domains, hindering its widespread applicability.
- Computational Bottlenecks: Agentic systems highly suffer from resource consumption and latency. Large-scale models (LLMs + vision models) require high GPU/TPU power for each task. Continuous inference loop (planning + acting + observing) consumes a large number of tokens, dozens of iterative LLM calls, and introduces round-trip delay, making sustained autonomy expensive at scale. Most organizations lack the infrastructure needed for high-volume deployment and often report that their systems are slow or costly.
- Ethical and Governance Deficits: Agentic systems require broad permissions across tools and access to sensitive data, which increases the risk of data leakage/ misuse and vulnerability to adversarial attack, creating unprecedented privacy risk. Furthermore, accountability gaps- who is responsible when an autonomous agent makes a harmful decision, combined with explainability deficits, bias, and regulatory uncertainty, are the main concern regarding the deployments of Agentic systems.

## 5 Applications of Agentic AI

While Agentic AI is a recently evolving field, various applications of this technology can already be found in the literature. Some of the applications in particular fields are given in the sub-sections below:

### *5.1 Human-Computer Interaction (HCI)*

In human-computer interaction (HCI), Agentic AI focuses on transparency, integrity, and adaptive collaboration. Integrity-based and contrastive explanations, particularly those aligned with users' mental models, increase trust and understanding [71,72]. Frameworks that preserve paradata enhance post-interaction transparency and traceability [73]. Transparency, even in undesired agent behavior, has been shown to increase user satisfaction and trust [74]. Cognitive agents that are user workload adaptive and capable of explaining themselves [75], as well as ethical training agents for individuals with autism spectrum disorder (ASD) [76], demonstrate that transparency supports effective, trustworthy human-AI teaming. Socially intelligent agents capable of conversation memory and topic-switching improve group chat dynamics [77], while spatial auditory perception systems enable realistic 3D interaction [78]. These systems further enhance Agentic AI's perceptual capabilities.

Beyond interaction, Agentic AI also shapes user behavior. AI-powered search agents support natural language tasks and information synthesis beyond keyword matching [4]. Agents can even shift intertemporal choice by compressing users' perception of future time, promoting long-term decision-making [79].

However, user controllability and transparency are demanded from proactive AI agents. Users desire to override capability and rationality [80] and often rely on personal confidence over AI accuracy [81]. User perception of AI is crucial, as users behave more unethically with service agents than enforcement agents, due to lowered accountability [82].

Across the reviewed studies, Agentic AI demonstrated strong performance in user-interaction environments, particularly when continuous feedback and personalization are needed. Empirical observations suggest that Agentic AI can achieve better performance through capabilities such as memory retention, contextual reasoning, goal-driven actions, and multi-step interaction planning. These systems excel at enhancing user trust and engagement. However, these effects are observed in structured environments, where interaction boundaries and objectives are clearly defined.

Although Generative AI systems can provide human-like responses, most remain prompt-driven and limited to short-context interactions. Agentic AI enables reasoning and action through iterative feedback loops, allowing systems to adapt dynamically to users over time. In HCI contexts, such features of Agentic AI are quite crucial since personalization and context retention are essential. Agentic AI thus moves beyond static user responses toward goal-driven human-AI interactions, enabling systems to anticipate user needs and dynamically adjust interaction strategies.

Despite these advancements, many limitations still exist that prevent the real-world deployment of Agentic AI in HCI. From an architectural perspective, maintaining persistent personal user models and long-term contextual memory remains a significant challenge, particularly in large-scale systems. Furthermore, continuous planning and action loops introduce latency and increase resource consumption, thus increasing the computational load. From an ethical perspective, concerns may arise regarding user manipulation, transparency, and controllability, especially in systems capable of influencing behavior and decision-making. Additionally, in certain human-AI interaction settings, user demand overrides capabilities [80] and relies on personal judgments [81], thus highlighting the need to balance autonomy with user control.

### 5.2 Healthcare

In healthcare, Agentic AI is primarily applied through conversational agents and autonomous diagnostic systems. LLM-based chatbots support mental health by interpreting patient moods and providing personalized conversations [83]. Some chatbot models using retrieval-augmented generation (RAG) and reinforcement learning offer more empathetic and context-aware responses, improving social connectivity among isolated patients [84,85]. Beyond mental health, agentic AI aids in diagnostics. Agents have been used to detect antimicrobial resistance [86], and the AutoOmics framework fully automates multi-omic data analysis without expert input [87]. In surgery, reinforcement learning agents deployed in vascular robotics enable autonomous catheter navigation and decision-making [88]. These applications show Agentic AI's growing role in delivering personalized care, accelerating diagnostics, and supporting complex clinical tasks.

Healthcare applications of Agentic AI can be broadly categorized into i) conversational systems for patients and ii) clinical decision-support and autonomous systems. The conversational agents primarily focus on emotional intelligence, personalization, and enhancing patient engagement in mental health applications. Such agents are easier to deploy and scale than decision-support AI systems. Decision-support, diagnostic, and surgical systems focus on precision and autonomy. Unlike conversational agents, these systems require more validation, reliability, and explainability to ensure clinical robustness.

Agentic AI in healthcare provides advantages over other AI systems, such as continuous monitoring and adaptation to patients and their issues. These features are significant in the healthcare context, where decisions are high-stakes and context-dependent. These systems also support multi-step workflows, such as diagnosis, treatment planning, and intervention, making them more goal-directed clinical support systems. Challenges such as integrating multi-modal data, maintaining long-term patient models, and increasing computational demands still remain. More importantly, in the healthcare context, the most significant challenges remain ethical and regulatory concerns. Concerns such as data security, patient privacy, and clinical safety pose significant adoption barriers. Trust and adoption also remain limited due to healthcare professionals' need for transparency and control over autonomous systems.

### 5.3 Education

Agentic AI is increasingly integrated into educational settings to support personalized and adaptive student learning. GenAI agents, combined with traditional scaffolding methods, can enhance students' learning and understanding of complex subjects like visual analytics. Rather than passively responding to queries, these agents guide learners through data-driven stories using specific prompts that allow for more in-depth engagement [89]. This hybrid approach outperformed both the standalone GenAI and traditional scaffolding, suggesting that context-aware cue prompts and structured teaching methods enhance learning outcomes. Further, the AI-agent-supported collaborative learning (AI-CL) framework [90] demonstrates how large language model-based agents can dynamically scaffold programming education. Compared to traditional computer-supported collaborative learning (CSCL) environments, the agents reduce cognitive load but significantly improve the students' performance, engagement, and self-efficacy.

Agentic AI plays two roles in the educational context: individual learning assistants and collaborative learning agents. Individual agents focus on personalized tutoring, adaptive feedback, and knowledge reinforcement, whereas collaborative agents emphasize group interaction, coordination, and shared problem-solving. Findings suggest that Agentic AI assists learners by enabling continuous feedback and contextual adaptation. It benefits students by reducing cognitive load, improving conceptual understanding, and increasing engagement when interacting with agentic systems.

One of the most significant challenges preventing the adoption of such systems in the education field is the risk of students' over-reliance on AI assistance [91]. The roles of educators and teachers need to be carefully redefined to ensure that such systems complement rather than replace human instruction. Other challenges include protecting individual student data privacy, ensuring equal access to AI-driven tools, and maintaining long-term student models.

### 5.4 Consumer Services and E-Commerce

In consumer-facing environments, Agentic AI is increasingly used to automate service delivery, marketing, and customer interaction. A key design factor is anthropomorphism, i.e., the implementation of customer service agents with human-like traits. Moderate anthropomorphism optimizes customer patience during failures, while extremes (too human or too robotic) reduce forgiveness, trust, and user acceptance [92,93]. In [94], it is reported that affiliative humor increases forgiveness following failures. In [95], the authors demonstrated that positive emotional expressions boost satisfaction in simple tasks but may reduce credibility in complex interactions. Beyond interaction style, trust is also shaped by user expectations. In [96], it was found that AI agent failures hurt customer loyalty more than interactive voice response (IVR) system failures, especially for highly involved consumers who expect intelligent behavior. In another E-commerce context, the work in [97] positioned Agentic AI as the future of marketing, managing strategic and operational roles autonomously. Reference [98] proposed embedding test-driven development into agent workflows, enabling real-time adaptation for retail resilience. Similarly, reference [99] introduced EcoptiAI, an autonomous retail system that automates pricing, inventory management, and customer support, significantly reducing costs and latency. Lastly, continuous learning also plays a role. The work in [100] presented a feedback-based agentic framework that evolves in real time, reducing reliance on manual updates and enhancing customer engagement.

In Figure 6, a sample framework is shown, which can be a suitable example of an Industry-level smart transportation framework using an AI agent and Agentic AI, showing the superiority of the agentic system. The studies revealed the use of Agentic systems to either assist with customer service or support decision-making and process optimization. Compared to other AI technologies, such as Generative AI, Agentic AI enables goal-driven, autonomous workflows, anticipates user needs, adapts to user requirements, and executes complex tasks, including management and decision-making. Customer interaction agents focus on personalization and enhancement of user experience. These features are quite valuable in e-commerce, where user behavior differs vastly and continuous adaptation is essential to maintain user engagement. Decision-support and optimization agents focus on backend optimization, including recommendation systems, resource allocation, and workflow automation, to increase operational efficiency and service performance.

Integration of such modern Agentic AI systems with already existing infrastructure and legacy systems remains a major challenge. Given user involvement and the handling of personal data, concerns about data privacy, user profiling, and algorithmic bias also arise in such consumer-seller environments. Excessive automation and AI involvement may also reduce user trust if transparency is not maintained [101].

### 5.5 Network and Cyber-Infrastructure

Agentic AI is increasingly used to automate network diagnostics, optimization, and threat response. Cognitive agents outperform traditional methods in fault detection [102], while LLM-based agents enhance autonomous network life cycle management with improved accuracy and efficiency [103]. In software-defined networking (SDN), agent integration supports service demand prediction and resource scheduling [104]. Despite these benefits, agentic systems also expand the attack surface, introducing new cybersecurity risks [? ]. Addressing this, reference [105] proposes Mobile AI Agents (MAAs) for 6G networks, that are autonomous, edge-level agents capable of real-time perception, reasoning, and action. Their framework includes multi-agent collaboration, semantic communication, and continual learning, enabling adaptive, user-centric networking with minimal human oversight. These advancements position Agentic AI as key to intelligent, resilient, and self-managing future networks, while underscoring the need for built-in security.

The applications of Agentic AI in network and cyber-infrastructure can be categorized into autonomous monitoring systems and multi-agent coordination systems. While the former focuses on detecting anomalies, optimizing performance, and maintaining network stability, the latter focuses on systems that involve coordination among multiple agents and distributed decision-making. These Agentic systems were shown to perform effectively in network environments that require scalability and autonomous decision-making by improving resource utilization and reducing the need for manual human intervention. However, most of these features were observed in environments where system parameters and operational constraints are predefined.

Generative AI systems are limited because they are only used for data analysis and do not control the infrastructure to achieve a specific goal. In contrast, agentic AI operates in a goal-directed manner, continuously observing network states, making decisions, and executing actions in real time. Certain limitations still remain, such as interoperability and integration of such agentic systems with existing infrastructures. Maintaining coordination among multiple agents in distributed environments also creates new challenges. Continuous monitoring and decision-making processes require significant resources and can introduce latency in large-scale deployments. Granting agentic systems full autonomy may increase the risks of attacks and unintended system behavior. Furthermore, integrating these technologies into network infrastructure may raise additional governance concerns.

### *5.6 Structured Taxonomy Linking Agentic Architecture, Learning, and Application*

The previous sub-sections delineated different agent architectures, learning strategies, and application domains. However, the selection of topologies and learning mechanisms for each application domain is not arbitrary. It must be strictly decided by analyzing operational factors of the target domain. For instance, HCI demands unified cohesion, parallel synergy of inputs, and low latency, which is only present in centralized star orchestration. While aggregating multimodal signals and generating responses, P2P presents unpredictability while interacting with neighbors, tree topology, and hybrid orchestration lacks in response-time and has protocol complexity, and chained workflow suffers from inflexibility. Similarly, for learning strategy, RL requires a large number of trial-and-error episodes, which will frustrate the users. Even though RAG is good at pulling static facts, it is not useful to teach an agent how to behave. By contrast, Meta-RL and Layered memory will be the best choice for a learning strategy as they offer rapid adaptability and solve agent-myopia directly by storing valuable insights. Similarly, other domains with fundamentally different constraints demand completely different architectures and learning strategies. A detailed synthesized taxonomy is presented in Table 7, which maps the structural and cognitive design of Agentic AI to real-world applications.

## 6 Adoption Factors of Agentic AI Usage

As the development of Agentic AI continues to expand, its potential for use across various domains, as mentioned above, increases. However, several factors influence the adoption and use of this technology in these domains. Researchers have discussed various factors that can explain its rapid development, adoption, and use.

**Table 7:** Synthesized agent topologies, learning strategies, and application domains.

| Application Domain | Optimal Agent Topology | Primary Learning Strategy | Analytical justification |
|---|---|---|---|
| Human-computer Interaction (HCI) | Centralized star Orchestration | Meta-RL + Layered memory | Allows the orchestrator to act as a centralized brain by aggregating multi-modal signals rapidly, and maintaining long-term user context. |
| Healthcare | Tree Topology (Hierarchical) | RAG + Meta-RL | Such a topology ensures safety protocols are met before clinical actions are taken, while RAG and Meta-RL enhance explainability and zero-hallucination fact retrieval. |
| Education | Hybrid orchestration or Chained Workflow | RAG + RL | Allows the system to interact with students in a peer-to-peer manner, and the reward signal enables the system to refine its teaching strategies. |
| Consumer services & E-commerce | Marketplace | Online Learning + RAG | Allow agents to bid for resources and rapidly adapt to consumer trends. |
| Network & Cyber-infrastructure | P2P Mesh | Distributed RL | Such topology and learning reduce single-point-of-failure bottlenecks and enable the system to detect distributed threats with minimal latency. |

Additionally, these factors are also proposed as system quality dimensions of Agentic AI. The idea is built on the theory of system quality dimensions in [106], which discusses the attributes of technology that impact users' behavioral beliefs and intentions regarding the technology. Therefore, these identified attributes of agentic AI can influence stakeholders' intentions to adopt and use the technology in personal or industrial settings. These factors are outlined as follows:

### *6.1 Simplified and Faster Decision Making*

AI-based systems foster enhanced decision-making. As decision-making occurs across various contexts and domains, AI agents provide intelligent assistance for complex tasks. Agentic AI aims to take it to the next level by simplifying and making it faster. In disaster management, Agentic AI simplifies decision-making by providing a near-global solution [107]. Moreover, since different agents act autonomously in the system, collective decision-making becomes faster and more optimized for the whole system.

Nevertheless, due to the efficiency of the decision-making, the system becomes optimized, resulting in increased resilience of various system components [98]. To make it even better, agents act autonomously with more situational awareness, which improves the AI agent's goal selection mechanism in cognitive architectures. Such a scenario also exists in digital twins and vehicular systems [73].

### *6.2 Trust and Reliance on the System*

This is one of the outcomes of enhanced decision-making. As AI agents learn from user interaction and input, their proactive nature and responsiveness create user reliance, specifically if they have a lower level of proficiency in a specific domain [108]. The reliance ultimately results in establishing trust in these agents as time passes. In the organizational context, as AI agents learn to adapt to the environment and context, it also helps to create trust in AI-based decision-making systems [109]. As user experience improves after collaborating with these agents, their increased confidence level later helps design Agentic AI systems. However, there are various aspects of decision-making apart from trust and reliance that are also discussed in extant literature, such as the moral identity that can influence the unethical behavior of AI agents, specifically in consumer decision-making [82]. Moreover, AI agents with their autonomous and enhanced human interaction enhance trust and reliance on Agentic AI systems.

### *6.3 Sustainability, Resource Optimization, and Operational Efficiency*

AI agents working autonomously on different tasks can contribute to resource optimization. In SDN, the agents perform dynamic resource scheduling in the network to provide quality service [104]. Therefore, this proactive nature of resource allocation and conflict resolution helps to enhance operational efficiency and quality of service. Resource management based on environmental and contextual needs is also one of the fascinating tasks of AI agents [110]. They facilitate continuous resource and facility management based on the environmental data inside the infrastructure. Smart infrastructures, such as buildings, network infrastructures, and other industries, can potentially increase the superior quality of service by utilizing Agentic AI's potential [108].

### *6.4 Enhanced Human AI Interaction and Seamless Service Provider*

Agentic AI facilitates 24/7 customer service with enhanced user interaction [111]. It also helps increase user engagement, provide personalized recommendations, and deliver real-time information to support informed decision-making [92]. Moreover, Agentic AI helps develop systems that handle information overload across various community groups, enhance user communication, foster an emotional companionship experience by independently organizing and managing conversation topics, and handle responses, etc. This type of seamless service aims to increase the efficiency in customer service, address the shortcomings of service providers, etc. [95]. The seamless service of AI agents in the education domain fosters an interactive and supportive learning environment for the students. The proactive nature of these agents facilitates instant and contextual responses, personalized explanations, which in the long run reduce the cognitive load of the learners [90]. Moreover, Agentic AI enhances customer engagement by providing personalized advertisements, tailored, real-time, and natural interaction [94].

### *6.5 Creativity, Innovation, and Increased Productivity*

The autonomous, independent operation and decision-making of Agentic AI enhance productivity across different domains. Lack of human intervention reduces the time, and learning from past experience accelerates scientific discovery and innovation [25]. Moreover, AI agents take distinct responsibilities separately to manage tasks effectively. These tasks are then performed with optimized resource planning and usage, which increases productivity [4].

### *6.6 Scalability and Handling Large Volume of Data*

Given the modular nature of Agentic AI, systems can be scalable and handle large volumes of data [112]. Archival data can be analyzed to predict better performance, develop more efficient comparative analysis, recommend new strategies for business development, optimize and utilize off-the-shelf ideas and solutions, etc. Similarly, data can be analyzed from different perspectives, and diverse analytics can be achieved to enhance the versatility [113]. For example, waste management, food consumption, and climate change data can be analyzed to comprehensively analyze environmental monitoring.

## 7 Synthesized Framework of Agentic AI Development and Adoption

The findings of this research work have contributed to the proposal of a comprehensive, synthesized framework (Figure 7) for the research and adoption of Agentic AI. The framework is based on the research work of Wixom and Todd [106] and it talks about the Object based attitudes (system quality such as simplified and faster decision, reliance, resource optimization and operational efficiency, scalability and data handing, seamless service provider) of Agentic AI enabled systems and how these qualities contribute to the behavioral beliefs (system satisfaction such as enhanced HCI, Trust and reliability). These behavioral beliefs have implications for behavioral intention (Agentic AI adoption, Agentic AI use). Wixom and Todd [106] discussed the object-based beliefs as attributes of technology, and behavioral beliefs are the possible effects when the technology is used. They investigated that the impacts of object-based beliefs on behavioral beliefs are mediated through the object-based attitude [114]. Besides this theory, another empirical investigation shows that the impact does have to be mediated; that is, object-based beliefs can have direct impacts on behavioral beliefs. Hence, the proposed framework figure shows a direct outcome rather than a mediated impact.

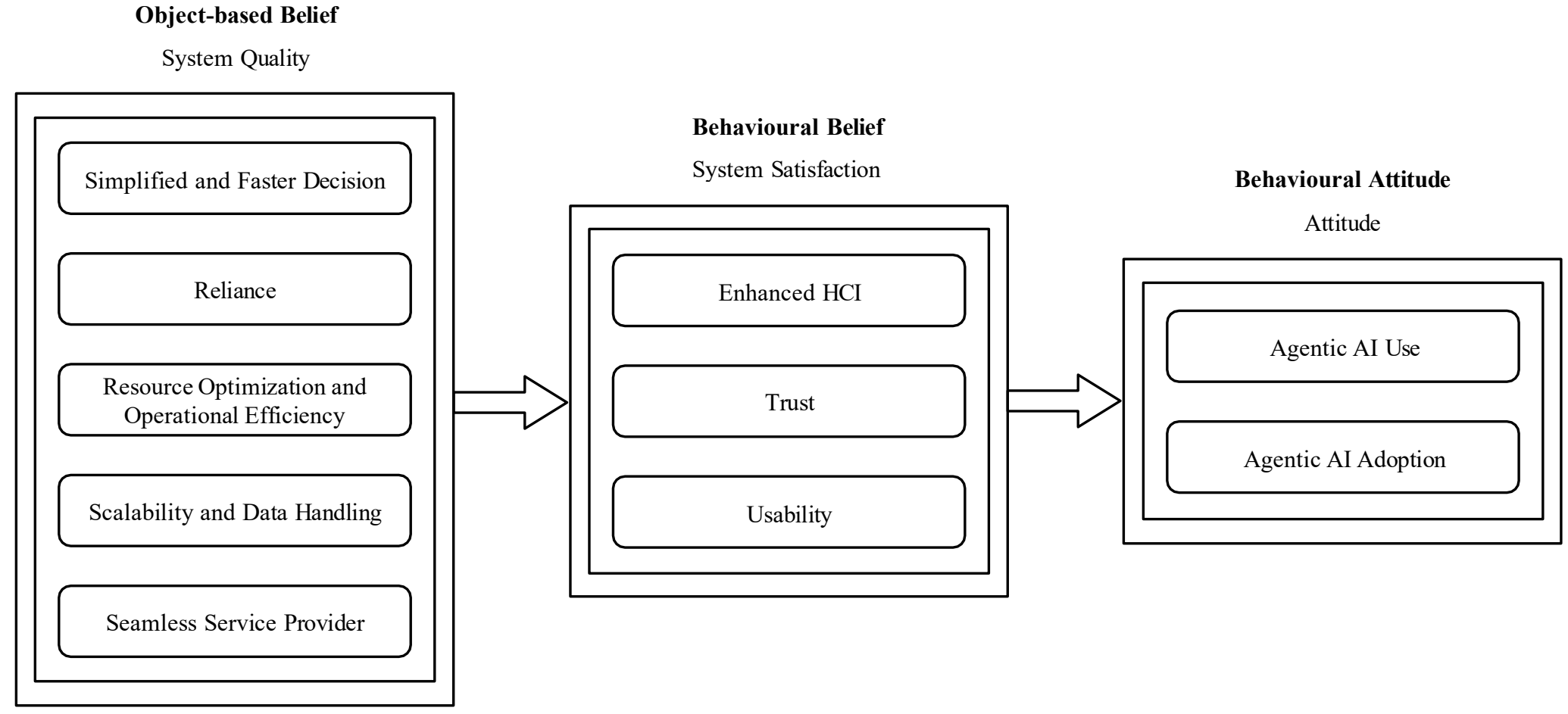

**Figure 7:** Synthesized Framework of Agentic AI development and adoption

## 8 Critical Analysis of Future Research Agendas

The SLR shows that agentic AI is emerging and offers significant scope for improvement. To identify limitations and future research directions, Alvesson and Jörgen Sandberg's problematization methodology has been adopted [115,116]. This framework fosters in generating critical standpoint by identifying limitations and challenges within existing literature, rather than simply identifying "gaps" (gap-spotting). Therefore, critical research directions are identified by analyzing "what" the current knowledge is and "how" it can be improved. The observation shows that the application domains of Agentic AI systems are either theoretical or not implemented on a large scale. Moreover, for widespread implementation, various other perspectives, such as frameworks, lay user issues, regulatory perspectives, etc., need to be considered. Based on the problematization approach, the whole observation was divided into 8 primary thematic categories. Furthermore, rather than simply pointing out gaps in the research findings, we have sought to articulate emerging research questions arising from unexplored areas. We then constructed them based on their potential significance to identify specific, feasible tentative RQs. The criteria for research gaps and future research directions are presented chronologically by significance. The significance is analyzed based on consensus from the literature and the field's development trend to identify the most urgent gaps and those worth exploring first. The categorized future research directions are presented in Tables 8-13 in a summarized format.

- **Ethical and Behavioral Considerations:** Research should explore how transparency influences human–agent collaboration in complex and long-term scenarios, with a focus on designing mechanisms that promote accountability, trust, and explainability [2,13,69,76–82,117]. Studies should also investigate extended human–AI interactions to better understand unethical behavior and its mitigation. Moreover, frameworks should be developed to address hermeneutic harm, manage emotional responses, and regulate reactive attitudes in AI-related incidents, thereby improving the ethical design and deployment of AI [117]. Tentative future research directions include exploring the impact of transparency on human-agent collaboration in complex scenarios, using qualitative (e.g., Interviews, focus groups) or quantitative (e.g., Experiments, surveys) methods[126–128]. Moreover, quantitative trust metrics can be developed through cross-sectional/longitudinal studies, along with co-design workshops/participatory design to implement and compare explainable AI (XAI) frameworks for visualizing agent decision trees [127] [128][129]. In order to address the ethical behavior-related research gaps, future research gaps can be: (1) develop behavioral audit frameworks to detect ethical violations in agentic systems using design science research/design thinking/participatory design approach, (2) investigate red-teaming protocols to detect emergent unethical behavior from misaligned reward functions. Addressing research gaps can help implement design, ethical, and responsible agentic AI across various domains, specifically for mission-critical/sensitive systems and high-stakes decision-making, such as finance, healthcare, transportation, etc
- **Data and Methodological Gaps:** After critically analyzing the existing literature, the data shows that there are research gaps in the availability of data that can help examine the impact of data imbalances and the lack of diversity on AI performance, etc. Additionally, based on the data analysis, there are research gaps that investigate the discrepancy between controlled testing and real-world performance [7,9,18–20,74,111–113]. Future research needs to focus on improving dataset diversity and scale to enhance AI generalizability, particularly in medical and social contexts [99]. Designing robust field studies and longitudinal evaluations will be essential for validating real-world applicability and long-term performance of Agentic AI. Additionally, methods to better organize and leverage fragmented domain knowledge [130] must be developed to support more effective AI system development and application. To address the research gap and to implement future research directions, cross-sectional/experimental research can be used, and AI agents can be deployed alongside human baselines to establish pre- and post-performance benchmarks[128]. Additionally, to develop failure taxonomies of real-world use cases that are not captured in a controlled test environment, action research and controlled experiments can be used [128]. To address real-world research gaps in dataset availability across domains, cross-domain,

**Table 8:** Critical Analysis of Research Gaps and Future Research Directions (Part 1)

| Criteria | Research Gap | Future Research Direction/Paths | Tentative RQs | Priority |
|---|---|---|---|---|
| **Ethical and Behavioral Considerations** [2,13,27,28,33,57,69,77–82,90,91,93,94,117] | Trust and Transparency: Challenges in quantifying the cognitive effects of trust and transparency. | (1) Explore the impact of transparency on human-agent collaboration in complex scenarios, Qualitative (Interviews, Focus Groups, etc.) or Quantitative (Experiments, Surveys, etc)/ research. (2) Develop quantitative trust metrics using cross-sectional/longitudinal studies. (3) Conduct co-design workshops/participatory design to implement and compare explainable AI (XAI) frameworks to visualize agent decision trees. | RQ1. In long-term human-agent collaboration, how do explainability mechanisms affect user trust as task complexity increases and agent behavior evolves? | • • • • • |
| | Ethical Behavior: Examines the drivers of unethical behavior and the impacts of AI design on ethical outcomes. | (1) Develop behavioral audit frameworks to detect ethical violations in Agentic systems using design science research/design thinking/participatory design approach. (2) Investigate red-teaming protocols to detect emergent unethical behavior from misaligned reward functions. | RQ2. In mission-critical systems (e.g., healthcare and legal advisory), how can behavioral auditing address unethical behavior? RQ3. How can constrained RL identify and ensure ethical Agentic task execution? | • • • • • |
| | Hermeneutic Harm: Investigates the nature and implications of hermeneutic harm in AI ethics. | Conduct systematic review and analyze the literature to construct a semantic gap based on hermeneutic harm taxonomies for Agentic AI. Develop reactive and policy-aligned algorithms to distinguish between benign and threat actors. | RQ4. How can hermeneutic harm taxonomies be utilized to detect and mitigate secondary harm? RQ5. How can reward shaping strategies be utilized to regulate reactive attitudes in AI-related incidents? | • • • • • |

**Table 9:** Critical Analysis of Research Gaps and Future Research Directions (Part 2)

| Criteria | Research Gap | Future Research Direction/Paths | Tentative RQs | Priority |
|---|---|---|---|---|
| **Data and Methodological Gaps** [7,9,18–20, 54,56,58,67,74,90, 91,112,113] | Data Limitations: Examines the impact of data imbalances and lack of diversity on AI performance. | (1) Construct cross-domain, multi-modal datasets across several domains (healthcare, financial, legal, educational, and industrial settings). (2) Apply weakly supervised and self-supervised augmentation for data-scarce scenarios. (3) Develop federated learning and differential privacy protocols for synthetic data, ensuring confidentiality and reliability. | RQ6. In medical diagnostics, how can cross-center, multi-modal datasets be constructed to address the impact of data distribution shifts on the generalizability of AI agents? RQ7. To construct cross-center, multi-modal datasets, how can federated learning and differential privacy frameworks be implemented to satisfy domain-specific data privacy and regulatory constraints(e.g., HIPAA, EU AI Act)? | • • • • ◦ |
| | Real-World Applicability: Investigates the disconnect between controlled testing and real-world performance. | (1) Use cross-sectional/experimental research and deploy AI agents alongside human baselines to prepare pre/post performance benchmarks. (2) Use Action Research/Controlled Research to develop failure taxonomies of real-world use cases that are not captured in a controlled test environment. (3) Implement tracking protocols to monitor concept drift and performance degradation. | RQ8. In industrial settings, how can longitudinal research be designed to detect drifts according to predefined thresholds and identify failure modes in real-world edge cases? | • • • • ◦ |
| | Fragmented Knowledge: Examines the effects of knowledge fragmentation on AI development. | (1) Construct ontology-based fragmented domain rules mapping fusion pipelines for cross-domain knowledge integration. (2) Build domain-specific knowledge graphs or databases and connect Agentic AI systems to enable context-aware AI reasoning | RQ9. In precision additive manufacturing and engineering design, how do knowledge graphs and RAG systems unify fragmented technical constraints to improve process planning and reduce AI Agent performance degradation? | • • • • ◦ |

**Table 10:** Critical Analysis of Research Gaps and Future Research Directions (Part 3)

| Criteria | Research Gap | Future Research Direction/Paths | Tentative RQs | Priority |
| --- | --- | --- | --- | --- |
| **Technical and Computational Gaps** [7,105,118] [9,12,24,37,54,56, 58,67,90,91] | Scalability and Efficiency: Investigates technical barriers to scalable and efficient AI deployment. | (1) Enhance real-time adaptability and computational efficiency, particularly for 6G networks and robotics. (2) Develop energy-efficient transformer variants and implement federated learning protocols for a scalable technical solution. | RQ10. Under dynamic, resource-constrained deployment conditions, how can real-time learning systems and energy-efficient architectures be developed to enhance Agentic AI performance in 6G networks and robotics? | •••◦◦ |
| | Goal Alignment: Examine issues with performance vs. goals. | Develop multi-object reward architectures using Pareto optimization and extract optimal reward parameters balancing efficiency, safety, and fairness. | RQ11. In complex, multi-step planning, what methods can be developed to optimize reward function parameterization and increase degrees of freedom in reinforcement learning-based AI agents? | •••◦◦ |
| | Explainability and Interpretability: Investigates the impact of poor interpretability on AI performance and trust. | Investigate explainability techniques (i.e. LIME, SHAP, Anchors, etc) using experimental studies involving necessary stakeholder. Advance explainability through techniques like Wasserstein Autoencoders and guided adaptive frameworks. | RQ12. In high-stakes Agentic AI systems (e.g., clinical or financial decisions), how can techniques such as Wasserstein Autoencoders and guided adaptive frameworks enhance their explainability compared to standard baseline approaches? | •••◦◦ |

**Table 11:** Critical Analysis of Research Gaps and Future Research Directions (Part 4)

| Criteria | Research Gap | Future Research Direction/Paths | Tentative RQs | Priority |
|---|---|---|---|---|
| **Regulatory and Governance** [2,25,69,81–85,90, 91,119] | Governance: Lack of a governance mechanism that addresses Agentic AI activities, data handling, communication protocols, security, and regulatory issues. | (i) conduct a systematic literature review to identify the actors of the governance framework, (ii) use design thinking/participatory design to design, develop, and evaluate a comprehensive governance framework, (iii)Design science research can be used to develop and rigorously evaluate a domain-specific governance framework. | RQ13. How can an Agentic AI governance framework be developed that addresses data handling, communication protocols, security, and regulatory compliance? RQ 14. How can an Agentic AI governance framework be designed and developed to handle mission-critical/sensitive domains such as healthcare, cybersecurity, finance, etc.? | • • • ◦ ◦ |
| | Governance framework for different Infrastructures: lack of governance and regulatory frameworks for different infrastructures and integration scenarios. Limited research on technical and operational challenges in expert system integration. | (1) Conduct systematic reviews and user studies (qualitative/exploratory) that identify the issues and actors of the governance framework in different infrastructure (2) Use co-design or participatory design for developing a governance framework and regulatory adoption guidelines for different infrastructures. | RQ15. In urban infrastructure planning, how can the Agentic AI governance framework address and optimize for data synchronization, data degradation, sensor failures, and other adversarial conditions? | • • • ◦ ◦ |

**Table 12:** Critical Analysis of Research Gaps and Future Research Directions (Part 5)

| Criteria | Research Gap | Future Research Direction/Paths | Tentative RQs | Priority |
|---|---|---|---|---|
| | Lack of socially interactive, domain-specific AI applications development | (1) Collect empirical data using controlled experiments or simulations. (2) Develop benchmarks (latency, social coherence, success rate, etc.) to train and evaluate algorithmic performance using participatory design/design science research. | RQ16. In a collaborative workspace, how can multi-modal algorithms be implemented to improve the real-time adaptability of socially interactive AI agents across varying levels of interaction complexity and population sizes? | ••◦◦◦ |
| **Human-AI Collaboration and Cognitive Bias** [66,73,76,77, 87,90,91,120] | Limited Human-AI interaction and cognitive bias: Lack of investigation of factors that impact the human-agent collaboration performance and cognitive bias. | (1) Design an embedded bias detection and reduction pipeline in human-in-the-loop (HITL) systems using a participatory design/design thinking approach. (2) Investigate trust calibration and cognitive bias in varying conditions in order to investigate the influencing factors using a cross-sectional study, experiments, and statistical analysis. | RQ17. In high-stakes environments (e.g., medical diagnosis support), investigate how and which factors impact human-agent collaboration performance and cognitive bias? | ••◦◦◦ |
| **Domain Specific Agentic AI Research** [73,90,91,118,119, 121–123] | Due to the recent development of Agentic AI, various domains such as transportation, aerospace, media and entertainment have not been widely explored. | Design domain-specific simulation environments, evaluation benchmarks, and cross-domain transferability validation protocols when AI agents migrate across application domains. | RQ18. In the transportation and aerospace domains, identify the specifications and the design requirements of applying Agentic AI? RQ19. In media and entertainment domains, what environments and evaluation benchmarks are necessary to validate Agentic AI agents before real-world deployment? | ••◦◦◦ |

**Table 13:** Critical Analysis of Research Gaps and Future Research Directions (Part 6)

| Criteria | Research Gap | Future Research Direction/Paths | Tentative RQs | Priority |
|---|---|---|---|---|
| **Integration with other Emerging Technologies** [74,81,97,119, 122–125] | Barriers to Integration: Lack of Investigation into socio-cultural and regulatory barriers to AI adoption across multiple domains and industry. | (i)Exploratory analysis using the qualitative research method to delve deeper into cultural and industry-specific strategies to overcome resistance to AI adoption. (ii)Use experimental research to conduct a quantitative investigation. (iii)Design thinking approach for strategy design, test, and development for agentic AI adoption. | RQ20. In cybersecurity and healthcare, what cultural and industry-specific strategies can be developed to facilitate the adoption of agentic AI? | • • • ◦ ◦ |
| | Lack of integration pipelines: Agentic AI with other technologies such as IoT, Wearable sensors, Quantum computing, and blockchain. | (i) Design IoT-Agentic data fusion pipelines to process heterogeneous sensory data. (ii) Develop quantum-computing/ blockchain/ wearable sensors or IoT-Agentic data fusion pipelines and establish protocols for cross-platform agentic AI integration | RQ21. How can an integration framework be developed using SW methodologies or Design Science Research for agentic AI? | • • ◦ ◦ ◦ |

multimodal datasets spanning several domains (healthcare, financial, legal, educational, and industrial settings) can be developed.

- **Technical and Computational Gaps:** The research should focus on the design of real-time learning systems and computationally efficient architectures capable of operating effectively in environments such as 6G networks [105] and robotics [118]. Developing scalable technical solutions that maintain adaptability at larger operational scales is crucial [7]. Further work should improve the parameterization of reward functions in reinforcement learning, advance explainability, and analyze AI agent interactions to promote greater interpretability. Similarly, explainability techniques (i.e., LIME [131], SHAP [132], Anchors [133], etc.) can be investigated using experimental studies involving necessary stakeholders. This research will help practitioners use appropriate techniques in real-world settings[128]. Explainability techniques can be contextualized for a specific domain to tailor them for industry-specific applications. Addressing these technical and computational research gaps will enable the development of scalable, efficient, and interpretable Agentic AI systems capable of operating reliably in real-world, resource-constrained, and high-demand environments. This will support the deployment of robust, explainable, and adaptive AI solutions across domains.
- **Regulatory and Governance:** Future research should be conducted on building comprehensive governance frameworks that include robust security inspection mechanisms and standardized communication protocols to reduce vulnerabilities in Agentic AI systems [109]. It is also important to design strategies tailored to industry-specific challenges to reduce resistance to adoption. Research should validate frameworks that facilitate regulatory-compliant adoption of Agentic AI and conduct clinical trials to meet the standards required in domains such as healthcare [25]. In order to address the research questions, these tentative techniques can be followed: (i) conduct a systematic literature review to identify the actors of the governance framework [16], (ii) use design thinking/participatory design to design, develop, and evaluate a comprehensive governance framework[134], (iii) design science research can be used to develop and rigorously evaluate a domain-specific governance framework[135], (iv) use co-design or participatory design for developing a governance framework and regulatory adoption guidelines for different infrastructures [135,136]. Addressing these regulatory and governance research gaps will facilitate the development of standardized, secure, and compliant Agentic AI frameworks, thereby enabling trustworthy adoption across regulated and high-risk domains.

- **Human-AI Collaboration and Cognitive Bias:** Improving human agent collaboration is one of the primary needs for future research. However, there is a lack of frameworks that aim to address these research gaps. The development of such interactive frameworks involves data collection across various environments and circumstances. Such can be used in the future for algorithm development, exploring strategies to strengthen collaboration, and conducting human studies to enhance dynamic capabilities and reduce bias [74]. Similarly, another exciting area can be to take the concept of Agentic AI and investigate tentative application use cases with a digital twin. Since Agentic AI operates autonomously and DT is one of the most effective ways to simulate real-world use cases, integrating both can be useful for that purpose. The findings from such a simulation can help design and develop an interactive, digital twin-based framework for human-agent collaboration across various domains. Practitioners can use the findings across industries such as transportation, healthcare, aviation, and construction, etc. However, several research gaps remain in the development of socially interactive, agentic AI. To address research gaps, human-in-the-loop experiments can be conducted. These experiments can be used to collect empirical data, develop benchmarks (latency, social coherence, success rate, etc.) to train and evaluate algorithms. Additionally, a participatory design/design thinking approach can be adopted to develop an embedded pipeline for bias detection and reduction in human-in-the-loop (HITL) systems. Another tentative direction can be to investigate trust calibration and cognitive bias across varying conditions to identify influencing factors using a cross-sectional study, experiments, and statistical analysis [126,129]. Addressing such research gaps will enable the design of human-centered, transparent, and bias-aware Agentic AI systems that enhance collaboration between humans and intelligent agents.
- **Domain Specific Agentic AI research:** Future work should conduct experiments with Agentic AI across underexplored domains such as transportation, healthcare, aerospace, finance, and social media [73]. This involves validating applications [121] through case studies and pilot projects while highlighting domain-specific challenges, design requirements, and metrics to ensure successful implementations.
  - **Agentic Smart Grid Systems:** Smart grid systems automatically predict the possible energy production and usage. This analysis is based on previous usage data and helps optimize energy distribution. However, this sector can be further improved using Agentic AI. As agents can work autonomously and collaborate to make better decisions, each agent can operate in separate sectors, such as energy production, distribution, and load balancing across various grids. Besides, these agents can be implemented across different distribution scenarios, such as urban, rural, and township areas. For example, agents can be used to predict future energy consumption and trigger a bump in energy production and a reduction in electricity generation at the plant. In this way, the power generation sector can also make use of its resources. Moreover, fault detection agents can help with maintenance and generate automated responses and reaction tickets. Autonomous coordination among these agents can result in better grid performance through optimal energy generation and distribution [122] [123].

  - **Gaming Industry:** The gaming industry has always been one of the most creative and dynamic application domains. Agentic AI can be used in different contexts here. In lieu of a centralized system, multiple agents can be designed to engage in different tasks. For example, one agent can be responsible for memory management, and another for persona management. The persona management agent needs to be dynamic in shaping player behavior. Similarly, another agent can coordinate with these two agents to design and adapt a coherent storyline. As the industry continues to develop more personalized, near-real-world gaming, Agentic AI can play a greater role in designing each storyline, which can later learn from player profiles to provide a more customized gaming experience to users. These techniques can not only develop gamers' experience but also enable developers to build more contextual, story-based multiplayer gaming where player agents can act like a digital twin of the real person behind the screen [119] [81] [124].
  - **Finance:** Autonomous agents can perform tasks separately, such as market analysis, stock market data analysis, real-time product demand and supply, etc. Besides, production analysis from the factory can also show potential delays and a lack of product delivery to customers. Agents can be helpful in analyzing and coordinating different tasks and optimizing the procedure. Additionally, agents can also use various APIs to communicate with different services and service providers in order to integrate them into the applications. For example, currency exchange rates can be analyzed alongside stock market analysis, company portfolio analysis, and real-time news data, and a risk analysis statement can be prepared for users. Users can make decisions based on the analysis. Therefore, Agentic AI has significant potential for use in the financial domain [74] [97].
  - **Media and Entertainment:** Agentic AI-based systems can act autonomously in the backend to analyze social media data and other media on any target platform. Based on the analysis, it can detect if a video or photo is fake and flag its user for future operations. Therefore, autonomous deepfake detection and handling can be performed with the advent of Agentic AI. Hence, such a system can be implemented in real life to study its performance and people's perceptions of it [125].
  - **Internet of Things:** In the future, Agentic AI can be used at the edge of IoT networks to improve decision-making efficiency by reducing latency and minimizing dependence on centralized cloud processing. Besides, AI agents can autonomously manage IoT devices by interpreting real-time and historical sensor data. This would make IoT systems more adaptive and efficient.

- **Integration with Emerging Technologies:** Research should identify and evaluate integration opportunities between Agentic AI and emerging/existing technologies such as IoT, blockchain, wearable sensors, quantum computing, and intelligent systems. Developing frameworks and methodologies for creating interoperable systems that combine multiple technologies will be essential to unlock new research and application possibilities [74,81,97,119,122–125]. Tentative research techniques include performing exploratory analysis using the qualitative research method to delve deeper into cultural and industry-specific strategies to overcome resistance to AI adoption; conducting experimental research to conduct a quantitative investigation; and using a design thinking approach for strategy design, testing, and development for agentic AI adoption [126–129,134,135].

## 9 Contribution to Research and Practice

The comprehensive overview of Agentic AI provides various perspectives, including application scenarios, domain-specific implementations, and research directions. The contributions of this work can be summarized as follows:

*Firstly,* this work critically and holistically outlines the historical perspectives and distinguishes between Agentic AI and other AI paradigms. To the best of our knowledge, a comprehensive description is yet to be outlined in scholarly contributions. Therefore, this work can be a very useful read and critical resource for both researchers and practitioners who would like to learn about the basics as well as the advances of Agentic AI.

*Secondly,* the work outlines the principles and components of Agentic AI in detail. The architecture and workflow of Agentic AI are illustrated in a way that is easily understandable to a reader with basic AI knowledge.

*Thirdly,* this work discusses potential real-world applications across various domains. The application not only discusses the empirical but also the proposed ones. Therefore, researchers and practitioners can see the latest applications, what has been implemented, and the proposed application areas. Hence, insights from this work can be useful to both AI researchers and practitioners in Agentic AI research and development.

*Fourthly,* this work critically analyzes the potential adoption factors of Agentic AI across various disciplines. These adoption factors reveal critical insights that can help practitioners and related stakeholders in their required feasibility study and identify room for improvement. Moreover, these adoption factors can serve as critical indicators and motivations for Agentic AI's real-world application. Agentic AI can help entrepreneurs, investors, and developers analyze and design optimized application scenarios.

*Lastly,* in this work, we critically evaluate the findings, applications, and research gaps. An investigation into the research gaps has identified promising future research avenues. These research avenues are categorized into domain-specific and application-specific directions. These future research directions are then also used to identify tentative future research questions. To the best of our knowledge, no other recent study on Agentic AI has adopted this approach to critically evaluate the future research avenues and directions.

## 10 Limitations

There are several limitations of this research. Firstly, this SLR used Scopus and Web of Science to search for scholarly publications. Publisher-specific libraries and Preprint servers (e.g., arXiv) have not been used as sources for article searches. Secondly, the proposed architecture (Figure 7) is not validated in this work, which can be an insightful future work for Agentic AI research.

## 11 Conclusion

Agentic AI, due to its emerging nature, needs to be investigated in terms of applicability and feasibility. Application across various domains should reveal crucial information regarding the research and development of this technology. Recent works show promising findings in terms of the multidisciplinary implementation of Agentic AI. Our investigation also reveals important factors to consider before adopting Agentic AI. However, the adoption factors discussed in this manuscript also offer practitioners an opportunity to adopt a more opportunistic perspective. Along with these implementation scenarios, this research critically discusses future research directions that are highly useful for both practitioners and researchers. This technology has significant potential for automating complex tasks. Therefore, Agentic AI will open the door to numerous applications across different service-providing scenarios. Hence, research on Agentic AI is timely and necessary for advancing AI.

**Acknowledgement:** During the preparation of this work, the authors used Grammarly Premium and ChatGPT in order to improve the sentence structure, coherence, readability, and language of the manuscript. After using this tool/service, the authors reviewed and edited the content as needed and take full responsibility for the content of the published article.

**Funding Statement:** The authors received no specific funding for this study.

**Author Contributions:** The authors confirm contribution to the paper as follows: Conceptualization, Research Design, Methodology, Data Collection, Analysis, Draft Writing, Validation, Visualization, Proof Reading: AKM Bahalul Haque; Conceptualization, Methodology, Data Analysis, Validation, Visualization, Draft Writing: Al Amin Islam Ridoy; Conceptualization, Data Collection, Analysis, Validation, Draft Writing, Proof Reading: Mohammad Rayhan; Conceptualization, Validation, Draft Writing, Proof Reading: Ivan Porres. All authors reviewed and approved the final version of the manuscript.

**Availability of Data and Materials:** Not applicable.

**Ethics Approval:** Not applicable, since this study does not involve humans or animals.

**Conflicts of Interest:** The authors declare no conflicts of interest.